\documentclass{article} % For LaTeX2e
\usepackage{iclr2027_conference,times}

\usepackage{amsmath,amsfonts,bm}

\def\eqref#1{equation~\ref{#1}}
\def\1{\bm{1}}

\DeclareMathAlphabet{\mathsfit}{\encodingdefault}{\sfdefault}{m}{sl}
\SetMathAlphabet{\mathsfit}{bold}{\encodingdefault}{\sfdefault}{bx}{n}

\usepackage{hyperref}
\usepackage{url}
\usepackage{amsmath}
\usepackage{amssymb}
\usepackage{amsthm}
\usepackage{graphicx}
\usepackage{wrapfig}
\usepackage{booktabs}
\usepackage{multirow}
\usepackage[normalem]{ulem}

\title{Compositional RL: Evaluating and Training LLMs for Collaborative Reasoning}
\author{
Shangbin Feng$^{1}$, Yulia Tsvetkov$^{1}$, Jan Kautz$^{2}$, David Acuna$^{2}$ \\
$^{1}$University of Washington \quad $^{2}$NVIDIA \\
\texttt{shangbin@cs.washington.edu}, \texttt{dacunamarrer@nvidia.com}
}

\iclrfinalcopy
\begin{document}

\maketitle

\begin{abstract}
Language models are often evaluated alone but increasingly deployed together, routed between, made to debate, ensembled, or collaboratively trained. For a model inside a system, what matters most is not how well it performs on its own, but how much the system gains from its presence. We call this a model's \emph{compositional strength} and formalize it using the Shapley value of a cooperative game over coalitions of a model pool, and use a model’s leave-one-out contribution to the full coalition as a tractable proxy, requiring $n+1$ rather than $2^n$ system evaluations. %and estimate it with leave-one-out marginals costing $n+1$ system runs rather than $2^n$. 
Evaluating individual vs.~compositional strengths across five model pools, twelve collaboration methods, and twelve datasets, we find them largely uncorrelated: in 12 of 15 (pool, method-family) settings, individual strength does not translate into compositional advantages, and the coupling is weakest among general-purpose models that practitioners are most likely to deploy. Standalone benchmarks are therefore often poor proxies for models' value in collaborative systems. We then propose \emph{compositional reinforcement learning} (C-RL), which trains a policy inside a live multi-model system and rewards the outcome produced by the system as a whole, rather than the policy’s correctness in isolation. The leave-one-out baseline in Shapley values cancels under a group policy gradient, so this needs no additional rollout cost beyond RL on the collaborative system's outcome and never requires running a coalition without the policy. Across three collaboration protocols and eight tasks spanning reasoning, coding, and scientific discovery, C-RL produces the strongest collaborator in 21 of 24 settings, lifting collaborative performance by $+4.0$ to $+5.4$ pts on average ($\sim$3x the gain of baselines), up to $+13.1$ pts. C-RL is the only method that never leaves the system worse off than the untrained policy, and carries over to unseen datasets, unseen collaboration protocols, and pools of frontier proprietary models.
\end{abstract}

\section{Introduction}

Language models are almost always evaluated alone. A model answers a benchmark, is scored, and lands on a leaderboard, all by itself \citep{liang2023holistic, chiang2024chatbot}. A growing share of deployed systems, however, do not use models alone. They route queries among several models \citep{ong2025routellm}, have models debate and revise \citep{du2023improving}, rank and fuse competing answers \citep{jiang2023llm}, or fine-tune a set of models to work together \citep{subramaniammultiagent,jiang2025sparta}. For a single participant in such a system, the quantity that matters is not how well it does the task on its own, but how much it contributes to the system. Standard evaluation does not capture this, and neither does standard training: typical post-training pipelines \citep{ouyang2022training, guo2025deepseek} optimize the same solo correctness that leaderboards reward, so a model's usefulness inside a system is neither measured nor optimized for. We call that quantity a model's \underline{\emph{compositional strength}}, and this paper asks how to evaluate it and how to train models for it.

\paragraph{Evaluating LLMs by compositional strength.} Inspired by cooperative game theory \citep{shapley1953value}, we treat a collaboration method on a task as a game whose coalitions are subsets of a model pool, and define a model's compositional strength as its Shapley value in that game (\S\ref{sec:eval}). Computing compositional strength exactly is, however, infeasible at scale: we estimate it by the leave-one-out (LOO) marginal, the drop in system performance when a model is removed, costing only $n+1$ runs, and tracking the exact value closely (Appendix~\ref{app:estimators}). If compositional strength is largely independent of individual capability, then existing evaluation and training efforts are inadequate for system-level intelligence.

\paragraph{Individual strength does not predict compositional strength.} Our extensive experiments across five model pools, twelve collaboration methods, and twelve datasets give a consistent answer. In 12 of the 15 (pool, method-family) correlation settings, individual and compositional strength bear no statistically significant relationship to each other, and in the three cells that do reach significance the correlation is weak ($|\rho| \le 0.29$) despite large samples. The decoupling is sharpest exactly where it matters most: among pools of capable general-purpose models no setting shows any significant relationship at all. A model's leaderboard standing thus says little about its value inside a system it joins.

\paragraph{Training LLMs for compositional strength.} We propose to train models with \emph{compositional reinforcement learning} (C-RL), which trains a policy inside a live collaborative system and rewards it with the system's outcome rather than with its solo correctness (\S\ref{sec:train}). The pool of frozen collaborators and the collaboration protocol together become the environment the policy acts in. We show training for compositional strength using LOO marginals actually doesn't require running an extra coalition that excludes the policy, so it needs no additional rollout cost beyond RL on the collaborative system's outcome.

\paragraph{C-RL reliably improves compositional strength.} We train Qwen~3.6 27B inside a pool of three frozen collaborators that are larger than the policy (gpt-oss-120b, kimi-k3, and nemotron-3.5-lightning), under three collaboration protocols and eight tasks spanning reasoning, scientific, multilingual, and pluralism. Across the 24 settings, C-RL\footnote{We will release data and code for both the evaluation and training studies upon acceptance.} yields the strongest collaborator in 21 and is the only method compared to five baselines that never leaves the system worse off than the untrained policy does. C-RL lifts collaborative performance by $+4.0$ to $+5.4$ points on average under each protocol, close to three times the gain of the strongest baseline, and by as much as $+13.1$ points on a single task, and RL with solo reward can sometimes make the policy individually better but the system measurably worse. The gains persist against a pool of frontier proprietary models (GPT-5.5, Gemini 3.5 Flash, Claude Opus 4.8) and transfer to datasets and collaboration protocols the policy never trained on, indicating C-RL teaches generalizable collaboration skills rather than fitting one setup. Together, these results chart a promising route toward preparing language models for system-level intelligence, a modular and collaborative AI future where what a model is worth is what it adds to the whole.

\section{Evaluating Language Models by Compositional Strength}
\label{sec:eval}

\subsection{Method}
\label{sec:eval-method}

Language models are typically evaluated and ranked by their \emph{individual strength}: how well a single model performs a task on its own. Though well suited to solitary deployment, it says nothing about a question that matters increasingly often in practice: when a model is deployed as one component of a larger collaborative system (multi-agent debate, model routing, ensembling, mixture-of-agents, or collaborative fine-tuning), how much does that model actually contribute to the system's success? We call this quantity a model's \emph{compositional strength}, and in this section we formalize it as a cooperative game and derive a tractable estimator for it.

\paragraph{Collaboration as a Cooperative Game.}

Let $\mathcal{N} = \{1, \dots, n\}$ denote a pool of candidate models. A \emph{coalition} is any subset $S \subseteq \mathcal{N}$ of models that collaborate together under a fixed collaboration method $m$ (e.g.\ a routing, debate, or ensembling protocol) to solve a fixed task $t$. We define the \emph{characteristic function}, the standard cooperative-game object assigning a scalar worth to every coalition:
\begin{equation}
v_{m,t}(S) \;=\; P_{m,t}(S) \;\in\; \mathbb{R},
\label{eq:value-function}
\end{equation}
where $P_{m,t}(S)$ is the performance (accuracy, pass rate, or any other task-specific score) obtained by running collaboration method $m$ on task $t$ using exactly the models in $S$, with $v_{m,t}(\varnothing)$ defined as a task-specific floor (e.g.\ zero or chance performance). Since $m$ and $t$ are fixed throughout this construction, we drop the subscripts and write $v(S)$ for the rest of the paper. The pair $(\mathcal{N}, v)$ is then a cooperative game: the worth of a group of models is simply how well the system they form performs. This presumes $v(S)$ depends only on \emph{which} models are in $S$ and not on any ordering among them, which is the case for almost all model collaboration algorithms, including every method we study; where a protocol needs an internal ordering, it derives one from the set itself.

Given this game, a model's compositional strength is its \emph{share} of the value the grand coalition $\mathcal{N}$ (the coalition of all models) creates. We use the Shapley value to define that share (see Appendix~\ref{app:shapley-axioms}), and estimate it in practice using LOO marginals.

\paragraph{The Shapley Value.}

The Shapley value $\phi_i(v)$ is model $i$'s marginal contribution, averaged over all $n!$ orderings in which the grand coalition is assembled one model at a time (more in Appendix~\ref{app:shapley-axioms}):
\begin{equation}
\phi_i(v) \;=\; \sum_{S \subseteq \mathcal{N} \setminus \{i\}} \frac{|S|!\,(n - |S| - 1)!}{n!} \Big[ v(S \cup \{i\}) - v(S) \Big].
\label{eq:shapley}
\end{equation}

\paragraph{Leave-One-Out as a Shapley Estimator.}
\label{sec:eval-loo}

Evaluating \eqref{eq:shapley} exactly requires $v(S)$ for all $2^n$ subsets of $\mathcal{N}$. When Shapley values are classically applied, the characteristic function is usually cheap, so enumeration is feasible; here a single $v(S)$ is an evaluation of a multi-model system, which puts exhaustive enumeration out of reach. We therefore estimate compositional strength with a \emph{leave-one-out} (LOO) marginal computed against the grand coalition:
\begin{equation}
\hat{\phi}_i(v) \;\triangleq\; v(\mathcal{N}) - v(\mathcal{N} \setminus \{i\}) \;=\; P_{m,t}(\mathcal{N}) - P_{m,t}(\mathcal{N} \setminus \{i\}).
\label{eq:loo}
\end{equation}
The hat distinguishes this estimate from the exact Shapley value $\phi_i(v)$ of \eqref{eq:shapley}; both are functions of the same game $v$, and we write $\hat{\phi}_i$ wherever it is clear from context. The compositional strength of model $i$ is the drop in the full system's performance when $i$ is removed: a positive value means $i$ was contributing, a value near zero means $i$ was redundant, and a negative value means $i$ was actively harmful to that particular collaboration.

The LOO marginal is not an unrelated heuristic: up to a factor $1/n$, it is exactly the $S = \mathcal{N}\setminus\{i\}$ term of the Shapley sum \eqref{eq:shapley},
\begin{equation}
\phi_i(v) \;=\; \underbrace{\tfrac{1}{n}\,\hat{\phi}_i(v)}_{|S| = n-1 \text{ term}} \;+\; \sum_{S \subsetneq \mathcal{N}\setminus\{i\}} \frac{|S|!\,(n-|S|-1)!}{n!}\Big[v(S \cup \{i\}) - v(S)\Big],
\label{eq:loo-decomposition}
\end{equation}
We confirm empirically (Appendix~\ref{app:estimators}) that this estimate is correlated with two more expensive ones over the same $v(\cdot)$, the Banzhaf value \citep{banzhaf1964weighted} and exact Shapley.
\paragraph{Computational cost.} Computing $\hat{\phi}_i$ for every $i \in \mathcal{N}$ via \eqref{eq:loo} requires only the $n+1$ coalition values $\{v(\mathcal{N})\} \cup \{v(\mathcal{N}\setminus\{i\}) : i \in \mathcal{N}\}$, linear in $n$, versus the $2^n$ coalitions required for exact Shapley. This is the reduction that makes a broad sweep across model pools, tasks, and collaboration methods feasible.

\subsection{Experiment Settings}
\label{sec:eval-setup}

We instantiate the framework of \S\ref{sec:eval-method} across a broad cross-product of model pools $\mathcal{N}$, tasks $t$, and collaboration methods $m$. We summarize each dimension below; full details of each model pool, every dataset, and the configuration of every collaboration method are given in Appendix~\ref{app:eval-setup}.

\paragraph{Model pools.} We evaluate five model pools spanning a wide range of scale and provenance: \emph{Specialized Expert Models} ($n=4$), domain-specialized $\sim$7B models fine-tuned respectively for general instruction-following, mathematics, coding, and long context; \emph{8-30B Open Models} ($n=7$), open-weight models in the tens-of-billions parameter range from several model families; \emph{$>$100B Open Models} ($n=4$), larger open-weight models; \emph{Proprietary API Models} ($n=3$), frontier proprietary systems served over a common API; and \emph{Diverse Academic Models} ($n=8$), a set of academically-trained models of heterogeneous origin. Each pool defines a separate cooperative game (\S\ref{sec:eval-method}); compositional strengths are computed independently within each pool.

\paragraph{Datasets.} We evaluate on 12 datasets spanning five broad task categories: \emph{QA} (ARC-Challenge \citep{clark2018think}, MMLU-Redux \citep{gema2025we}), \emph{Reasoning} (GPQA-Diamond \citep{rein2024gpqa}, BBH \citep{suzgun2023challenging}, TheoremQA \citep{chen2023theoremqa}), \emph{Coding} (HumanEval \citep{chen2021codex}, MBPP \citep{austin2021program}), \emph{Pluralism \& Safety} (CulturalBench-Hard \citep{chiu2024culturalbench}, CocoNot \citep{brahman2024art}), and \emph{Science} (SciRIFF \citep{wadden2025sciriff}, KernelBench \citep{ouyang2025kernelbench}, AssayBench \citep{debrouwer2026assaybench}).

\paragraph{Collaboration methods.} We evaluate 12 collaboration methods in three families that reflect qualitatively different modes of composition. \emph{Routing} methods dispatch each query or part of the query to one model from the pool: Prompt Routing \citep{frick2025prompt}, Trained Router \citep{ong2025routellm}, Switch Generation \citep{feng2025don}, and Graph Router \citep{fenggraphrouter}. \emph{Agentic} methods have several models interact and exchange generated texts before a final answer is produced: Multiagent Debate \citep{du2023improving}, Multiagent Feedback \citep{feng2024don}, Heterogeneous Swarms \citep{feng2025heterogeneous}, and SLM-MUX \citep{wang2026slm}. \emph{Training} methods involves training components or the models themselves for enhanced collaboration: LLM-Blender \citep{jiang2023llm}, AggLM \citep{zhao2025majority}, Multiagent Finetuning \citep{subramaniammultiagent}, and Sparta Alignment \citep{jiang2025sparta}. All methods are implemented within MoCo toolkit \citep{feng2026moco}.\footnote{\url{https://github.com/BunsenFeng/model_collaboration}}

\paragraph{Protocol.} For each valid (pool, dataset, method) cell we compute the $n+1$ coalition values required by \eqref{eq:loo} and derive each model's compositional strength $\hat{\phi}_i$. We additionally measure each model's \emph{individual strength}, its standalone score on the same dataset, by running it alone through the same evaluation pipeline. The result is a large matrix of paired (individual strength, compositional strength) observations, indexed by model, dataset, and collaboration method.

\subsection{Results}
\label{sec:eval-results}

\begin{figure}[t]
\begin{center}
\vspace*{-20pt}
\includegraphics[width=0.9\linewidth]{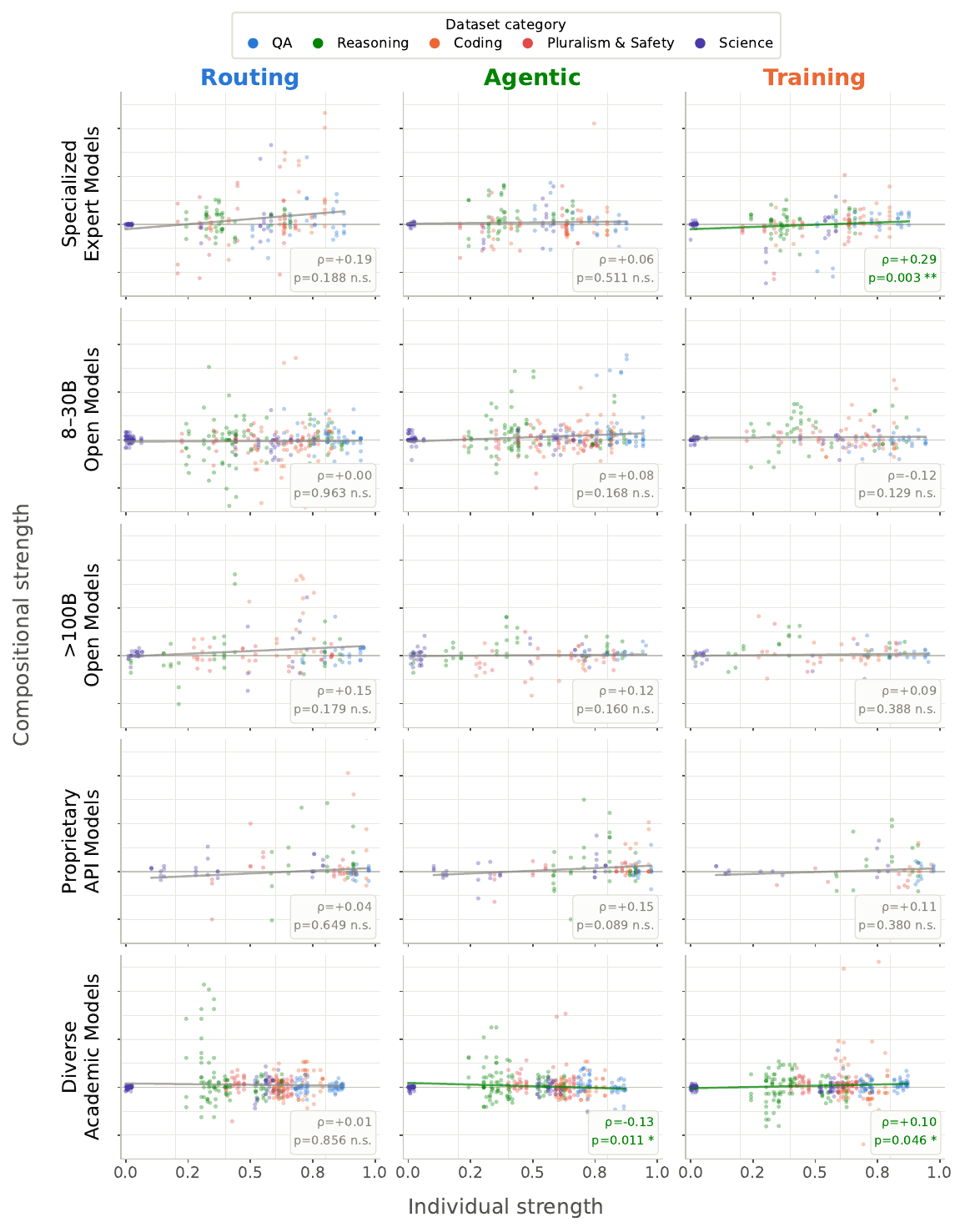}
\end{center}
\vspace*{-20pt}
\caption{Individual strength (solo score) vs.\ compositional strength (leave-one-out Shapley value in a collaborative system, \S\ref{sec:eval-method}), broken down by model pool (rows) and collaboration-method family (columns). Each point is one model evaluated on one dataset under one collaboration method; color denotes the dataset's task category. $\rho$ is the Spearman correlation with a permutation-test $p$-value ($10{,}000$ permutations within model). \textbf{Individual and compositional strength are largely uncorrelated}: in 12 of 15 (pool, method) panels the two show no significant relationship, visible from each panel's near-flat fitted line and its printed $\rho$ ($|\rho| \le 0.29$ throughout) and $p$-value (marked n.s.\ when $p>0.05$); the decoupling is sharpest among pools of general-purpose models, namely $>$100B Open Models and Proprietary Models (rows 3-4), where all six panels are n.s.}
\vspace*{-10pt}
\label{fig:overview-grid}
\end{figure}

\textbf{Collaboration algorithms are effective.} We first confirm that collaboration protocols improve performance (mean $+11.5$, 95\% CI $[+8.7, +14.7]$; details in Appendix~\ref{app:collab-effective}) across diverse (model pool, task) settings. The question now turns to correlation between individual strength ($v(\{i\})$, model $i$ acting alone) and the level of helpfulness, i.e., compositional strength ($\hat{\phi}_i$, \S\ref{sec:eval-loo}).

\paragraph{Individual and compositional strength are not correlated.} Across the full $5 \times 3$ grid of model pools and collaboration-method families (Figure~\ref{fig:overview-grid}), 12 of the 15 cells show no statistically significant relationship between a model's individual strength and its compositional strength. The remaining three cells reach significance only because of the large sample size in each cell; the correlations themselves are weak ($|\rho| \le 0.29$) and the fitted trend lines are correspondingly flat. A model's standalone benchmark score is therefore mostly uninformative about how much it contributes to the collaborative systems it joins.

\paragraph{The weak signal that does exist favors training-based collaboration.} Two of the three significant cells occur in the training-based column (Specialized Expert Models, $\rho=+0.29$; Diverse Academic Models, $\rho=+0.10$), where components or these models themselves are trained to work together. The routing column, in contrast, shows no significant relationship in any of the five pools. The pattern suggests that deeper levels of integration leave slightly more room for individual strength to matter. However, the effect is small even where it narrowly reaches significance, and it is absent from routing and agentic collaboration, which are more popular in day-to-day practices due to their simplicity and lower engineering overhead. For the collaborative systems practitioners are most likely to build, individual strength remains uninformative.

\paragraph{Larger and more capable model pools show the least coupling.} The three pools composed of larger or more capable models (8-30B Open Models, $>$100B Open Models, and Proprietary API Models) show \emph{no} significant correlation in any of the nine (pool, method-family) cells they span. All three significant cells instead occur in the two pools of specialization and diversity: Specialized Expert Models (narrowly fine-tuned $\sim$7B models) and Diverse Academic Models (a heterogeneous, non-frontier set). For capable general-purpose models, the pools practitioners are most likely to deploy, individual benchmarks carry no signal about compositional strength: this suggests that when using state-of-the-art models in collaboration, evaluating their synergy and complementarity is critical (\S\ref{sec:eval-analysis}).

\subsection{Analysis}
\label{sec:eval-analysis}

Now that individual and compositional strength are decoupled empirically, we study two complementary questions: which collaboration methods tend to produce the strongest collaborative systems in the first place, and where, across model pools and task domains, does collaboration help most?

\begin{figure}[t]
\begin{center}
\includegraphics[width=\linewidth]{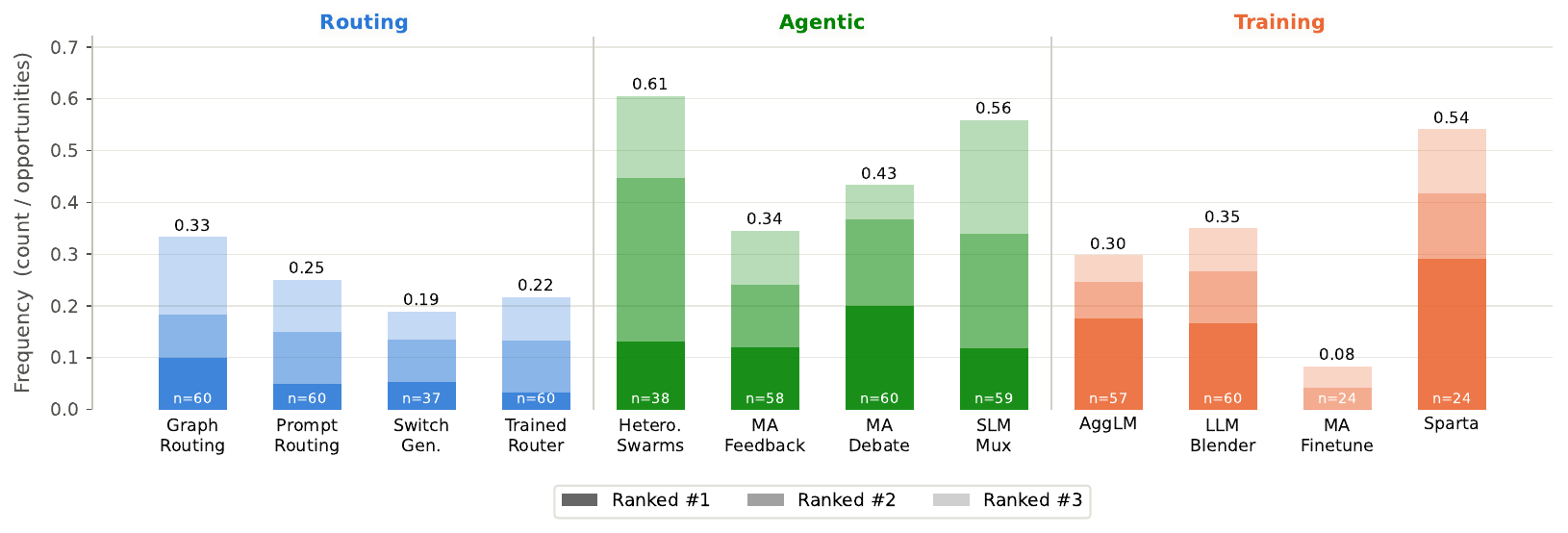}
\end{center}
\caption{Frequency when each of the 12 collaboration methods achieves the best (\#1), second-best (\#2), or third-best (\#3) full-coalition performance $P_{m,t}(\mathcal{N})$, out of all (pool, dataset) pairs in which it was evaluated ($n$ = number of such pairs). Ties share a rank.}
\label{fig:method-winrate}
\end{figure}

\begin{wrapfigure}[18]{r}{0.45\linewidth}
\centering
\vspace{-\baselineskip}
\includegraphics[width=\linewidth]{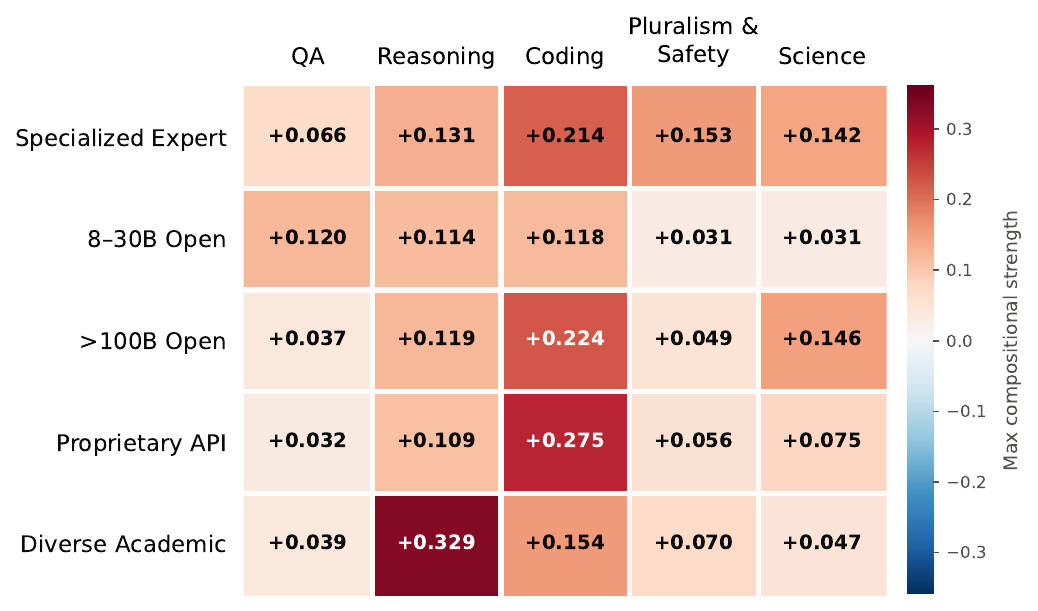}
\caption{Best compositional strength achieved by collaboration methods, for each (model pool, dataset category). Each cell reports $\max_{m,t \in \text{category}} \bar\phi(m,t)$, where $\bar\phi(m,t)$ is the mean compositional strength across models for $m$, $t$.}
\label{fig:domain-pool-heatmap}
\end{wrapfigure}

\paragraph{A small number of methods account for most top finishes.} Figure~\ref{fig:method-winrate} ranks each collaboration method by how often it achieves top-3 full-coalition performance across all (pool, dataset) pairs. The distribution is far from uniform: Sparta Alignment \citep{jiang2025sparta} and Multiagent Debate \citep{du2023improving} finish \#1 most often among the 12 methods, with by far the largest \#1 shares of any method, and together with LLM-Blender \citep{jiang2023llm} and AggLM \citep{zhao2025majority} they account for a disproportionate share of top-3 outcomes.

This complements results in \S\ref{sec:eval-results}: \uline{while \emph{which model} you add to a coalition is largely uninformative from its solo score, \emph{which collaboration method} you use is highly informative.} The agentic methods of collaboration focusing on multiple rounds of inference-time communication therefore strike a nice balance between engineering ease and strong performance, and we recommend them as first choices and future research on compositional strength in those systems.

\paragraph{Collaboration helps unevenly across task domains} Figure~\ref{fig:domain-pool-heatmap} aggregates the best achieved compositional strength for every (pool, domain) combination. Coding stands out as the domain where collaboration helps the most consistently: it is the strongest domain for three of the five pools (Specialized Expert, $>$100B Open, Proprietary API) and is never the weakest domain for any pool. QA, by contrast, is where collaboration helps the least: it is the weakest domain for four of the five pools, with the sole exception of the 8-30B Open pool. Reasoning shows a large but concentrated effect: Diverse Academic Models benefits from collaboration on Reasoning tasks more than any other single (pool, domain) cell in the grid, a peak not shared by the other four pools. Taken together with the win-rate result above, this indicates that the payoff from collaboration is governed far more by \emph{what kind of task} is being solved, \emph{which method} is used, and in some cases the specific composition of the pool, than by any single model's individual competence. This reinforces why compositional strength cannot be predicted from a model's individual benchmark score alone. A finer-grained, per-dataset (rather than per-domain) version of this analysis is provided in Appendix~\ref{app:gain-consistency}.

\section{Training for Compositional Strength}
\label{sec:train}

\subsection{Method}
\label{sec:train-method}

Section~\ref{sec:eval} establishes that compositional strength is a distinct property mostly uncorrelated with individual strength, the usual focus of existing evaluation and training efforts. We propose to train models for compositional strength directly rather than hope for it as a byproduct of individual-capability training, through \emph{compositional reinforcement learning} (C-RL).

\paragraph{Objective.} We denote a model collaboration method $m$, a task $t$, a trainable model as the policy $\pi_\theta$, and fixed collaborator models $\mathcal{C} = \mathcal{N}\setminus\{\pi_\theta\}$. We use $R$ for the task's reward function and $y_{\mathrm{sys}} \sim m(\cdot \mid q, S)$ as the answer the collaborative system produces on question $q$ from model coalition $S$, so that system performance $P_{m,t}(S) = \mathbb{E}_{q \sim t,\, y_{\mathrm{sys}} \sim m(\cdot \mid q, S)}\big[R(q, y_{\mathrm{sys}})\big]$. The pool of collaborators and the model collaboration algorithm $m$ become the \emph{environment} the policy acts in: training a model for compositional strength is thus reinforcement learning grounded in interaction with an environment of other LLMs. Taking the compositional strength of \eqref{eq:loo} as the quantity to maximize gives
\begin{equation}
\max_\theta \; \hat{\phi}_{\pi_\theta} \;=\; \max_\theta \Big[ P_{m,t}\big(\mathcal{C} \cup \{\pi_\theta\}\big) - P_{m,t}(\mathcal{C}) \Big].
\label{eq:crl-objective}
\end{equation}
To turn \eqref{eq:crl-objective} into a per-sample training signal for policy-gradient methods, the natural approach is to reward each sample by its own LOO marginal,
\begin{equation}
r(q, y_{\mathrm{sys}}) \;=\; R(q, y_{\mathrm{sys}}) - b_q, \qquad b_q \;=\; \mathbb{E}_{y_{\mathrm{sys}} \sim m(\cdot \mid q, \mathcal{C})}\big[R(q, y_{\mathrm{sys}})\big],
\label{eq:per-sample-loo-reward}
\end{equation}
subtracting the collaborators-only score $b_q$ on question $q$ from the sample's raw reward. Computed this way, $b_q$ requires an actual extra rollout of $\mathcal{C}$ alone for every question, doubling the number of system rollouts training would need.

We show this extra rollout is unnecessary. Group-relative policy-gradient methods already form the advantage of a sample by subtracting the mean reward over samples drawn for the same question. Letting $y^{(1)}, \dots, y^{(K)}$ be the system answers from $K$ sampled rollouts on $q$ with the policy plugged in, the advantage of the $j$-th sample, computed with the LOO-marginal reward $R(q, y^{(j)}) - b_q$, is
\begin{equation}
A^{(j)} \;=\; \Big[R\big(q, y^{(j)}\big) - b_q\Big] \;-\; \frac{1}{K}\sum_{l=1}^{K} \Big[R\big(q, y^{(l)}\big) - b_q\Big] \;=\; R\big(q, y^{(j)}\big) \;-\; \frac{1}{K}\sum_{l=1}^{K} R\big(q, y^{(l)}\big),
\label{eq:baseline-cancels}
\end{equation}
so $b_q$ cancels out entirely: group-relative advantage recovers exactly the LOO-marginal advantage, without ever running $\mathcal{C}$ alone to compute it. As a result, we can train directly on the system reward
\begin{equation}
\max_\theta \; \mathbb{E}_{q \sim t,\; y_{\mathrm{sys}} \sim m(\cdot \mid q,\, \mathcal{C} \cup \{\pi_\theta\})}\big[\, R\big(q, y_{\mathrm{sys}}\big) \,\big],
\label{eq:crl-reward}
\end{equation}
the expectation running over both the question draw and the policy and collaborators' stochastic generations, under a collaboration protocol $m$. Unlike a general Shapley or Banzhaf estimator, which would sample coalitions $S \subseteq \mathcal{N}$ to average over, C-RL always evaluates the one fixed grand coalition $\mathcal{C} \cup \{\pi_\theta\}$: this is exactly the leave-one-out simplification of \S\ref{sec:eval-loo}, and is what keeps every rollout to a single, cheap coalition rather than the many coalitions a general estimator would require. We can pin both $\mathcal{C}$ and $t$ to train a specialist policy, or unfreeze either/both by sampling collaborator sets/methods from a larger population or drawing tasks from a distribution.

\paragraph{The reward is the system's output, not the policy's own.} The C-RL objective indicates that the policy is never rewarded for being individually correct in solving the whole problem, but for how much its participation moves $y_{\mathrm{sys}}$ and improves system performance. A rollout in which the policy's own answer is wrong can still carry full reward if the system answered correctly (e.g., providing useful error patterns for other models to avoid), and a rollout in which the policy is right can carry zero reward if the system was wrong (e.g., not explaining the reasoning clear enough for other models to trust). This is the same distinction Section~\ref{sec:eval} shows individual benchmarks cannot see, now formalized as a training signal in C-RL.

\paragraph{Credit assignment across roles and turns.} A collaboration protocol may query the policy several times per question, and in several distinct roles. Let $\tau_1, \dots, \tau_L$ denote the policy's generations within one execution of the protocol on question $q$, where $L$ is itself a random quantity: a protocol may invoke the policy a different number of times from one execution to the next, depending on routing decisions, early termination, or degenerate outputs. Since $R$ scores only the final system output, we assign the same scalar to every generation in that execution,
\begin{equation}
r(\tau_\ell) \;=\; R\big(q, y_{\mathrm{sys}}\big) \quad \text{for all } \ell = 1,\dots,L,
\label{eq:turn-credit}
\end{equation}
so that each generation is credited with the system-level outcome it helped produce. Future work could explore differentiated credit assignment for diverse policy steps, but in this work we choose not to add this complexity while also finding that uniform assignment already works empirically.

\paragraph{Group-relative advantage over variable-length rollouts.} We optimize \eqref{eq:crl-reward} with a group-relative policy-gradient method \citep{shao2024deepseekmath}, which estimates the advantage of a generation by comparing its reward against a baseline formed from other samples for the same question, avoiding a learned value model. Sampling $K$ independent rollouts of the protocol for each of $N$ questions per step, and denote $\mathcal{T}_q$ as the set of all policy generations produced across the $K$ rollouts for question $q$, the advantage of generation $\tau \in \mathcal{T}_q$ is
\begin{equation}
A(\tau) \;=\; \frac{r(\tau) - \mu_q}{\sigma_q}, \qquad \mu_q = \frac{1}{|\mathcal{T}_q|}\sum_{\tau' \in \mathcal{T}_q} r(\tau'), \qquad \sigma_q = \mathrm{std}\big(\{r(\tau')\}_{\tau' \in \mathcal{T}_q}\big),
\label{eq:group-advantage}
\end{equation}
this is the mean-subtraction that produced the cancellation in \eqref{eq:baseline-cancels}. The one difference from standard GRPO is that the group is formed \emph{per question}, over the pooled generations of the policy model from that question's $K$ rollouts, rather than over a fixed number of samples per prompt. Because $L$, how many times the policy is called in a rollout, varies from one rollout to the next, $|\mathcal{T}_q|$ is not a fixed multiple of $K$, so the group size won't be fixed. Both $\mu_q$ and $\sigma_q$ are treated as fixed statistics of the sampled group, with no gradient flowing through them: dividing by $\sigma_q$ rescales every term in the group by the same positive constant, so it does not reintroduce $b_q$ or disturb the cancellation of \eqref{eq:baseline-cancels}, only the scale of the learning signal for that question. Because the loss averages uniformly over every policy-generation token in the batch rather than first normalizing by question, a question whose rollouts call the policy more often (larger $|\mathcal{T}_q|$) contributes proportionally more gradient signal in that step. The policy is then updated with the usual clipped policy-gradient objective \citep{schulman2017proximal} using $A(\tau)$, with the loss applied over the response tokens of each generation of the policy model.

\paragraph{Practical requirements.} Two things set this apart from standard RL post-training. First, the environment is a whole multi-model  collaborative system rather than a single-model rollout. It contains participants the trainer does not own, in our experiments including both open models much larger than the policy and closed-weight models served behind APIs. None of \eqref{eq:crl-objective} through \eqref{eq:group-advantage} requires differentiating through any of it; only the scalar $R(q, y_{\mathrm{sys}})$ crosses that boundary between policy and environment. Second, a rollout is not the product of a single model. The intermediate reasoning and final response include both the policy's own generations as well as collaborators' contributions, governed by the collaboration method $m$, yet a policy-gradient update is only meaningful on tokens the policy actually sampled. Applying the loss to collaborator text would teacher-force the policy on another model's distribution, which is off-policy and optimizes something other than \eqref{eq:crl-reward}. We therefore attribute different parts of a collaborative rollout to the model that produced it and train only on the policy's own generations, under the log-probabilities with which they were sampled. Appendix~\ref{app:crl-impl} describes how this attribution is implemented.

\subsection{Experiment Settings}
\label{sec:train-setup}

\paragraph{Policy and collaborators.} The trainable policy is Qwen~3.6 27B, full fine-tuning. Its collaborators are three fixed models, all larger than the policy model: gpt-oss-120b, kimi-k3, and nemotron-3.5-lightning, none of which is updated during training. Holding them frozen is what makes the baseline cancellation in \eqref{eq:baseline-cancels} apply. We vary the model settings in \S\ref{sec:train-analysis}.

\paragraph{Collaboration environments.} We train under three collaboration methods, chosen to span distinct modes of composition. LLM-Blender \citep{jiang2023llm} has every model produce a candidate answer, a ranker score the candidates, and a fuser synthesize the top-ranked ones into the system's answer; the policy is credited under \eqref{eq:turn-credit} for diverse roles. Multiagent Debate \citep{du2023improving} has models answer independently and then revise over successive rounds based on each other's responses. SLM-MUX \citep{wang2026slm} selects and runs several models in parallel and produces a final answer based on their outputs. Each defines a different collaborative environment in the sense of \S\ref{sec:train-method}, while we explore generalization to held-out collaboration algorithms in \S\ref{sec:train-analysis}.

\paragraph{Tasks.} We evaluate on eight tasks spanning QA, reasoning, and science: GPQA-Diamond \citep{rein2024gpqa}, graduate-level science questions; TheoremQA \citep{chen2023theoremqa}, theorem-driven mathematical reasoning; MMLU-ProX \citep{xuan2025mmlu}, multilingual multiple-choice questions across 20 languages; CulturalBench-Hard \citep{chiu2024culturalbench}, cultural reasoning and knowledge; ChemBench \citep{Mirza2025}, chemistry questions across nine subfields; KernelBench \citep{ouyang2025kernelbench}, rewriting PyTorch operators as faster custom CUDA kernels; SMDD-Bench \citep{han2026smddbench}, small-molecule drug design scored by executed constraint checks; and AssayBench \citep{debrouwer2026assaybench}, CRISPR genetic-screen gene ranking scored against real screen data. The first five are scored by matching the model's answer against a reference, either a multiple-choice option or an exact-match string; the next two are scored by executing the model's output in a scientific environment and measuring the result; and AssayBench is scored by ranking correlation against a precomputed relevance table, with no execution step. \S\ref{sec:train-analysis} explores generalization to held-out datasets.

\paragraph{Baselines.} We compare C-RL against five baselines, grouped by what they optimize. \emph{Base} is the untrained policy. \emph{Individual-reward RL} applies GRPO \citep{shao2024deepseekmath} and DAPO \citep{yu2026dapo} to the policy's own solo correctness. \emph{Collaboration-aware baselines}: \emph{prompt nudge} leaves the policy untrained and instead informs it in its system prompt how the collaborative system it is in works, and \emph{collaborative RAFT} \citep{dong2023raft, feng2026single} runs the real collaboration pipeline with the untrained policy, keeps the transcripts whose collaborative outcome scored well, and fine-tunes on them. These comparisons are designed to establish the effectiveness of training for compositional strength and optimizing it with reinforcement learning in C-RL.

\paragraph{Evaluation.} Every method is scored in two modes on the test set: \emph{individually}, with the trained policy model answering alone, and \emph{collaboratively}, with the trained policy inside the full collaborative environment. Checkpoints are selected on a held-out development split scored during training.

\subsection{Results}
\label{sec:train-results}

\begin{table}[t]
\centering
\footnotesize
\setlength{\tabcolsep}{-0.5pt}
\vspace*{-20pt}
\resizebox{0.8\textwidth}{!}{%
\begin{tabular*}{\linewidth}{@{\extracolsep{\fill}}l cccccccc c@{}}
\toprule
Method & GPQA-dia & Theorem & KernelB & SMDD & AssayB & ChemB & Culture-H & MMLU-PX & Avg. \\
\midrule
\multicolumn{10}{@{}l}{\emph{Collaboration protocol: LLM-Blender}} \\
  Base & .818 & .613 & .144 & .635 & .035 & .720 & .806 & .804 & .572 \\
  GRPO-indiv & .849 & .647 & .157 & .635 & .033 & .732 & .825 & .822 & .587 \\
  DAPO-indiv & \underline{.859} & \underline{.657} & .129 & .635 & \underline{.041} & \underline{.750} & .819 & .812 & \underline{.588} \\
  prompt-collab & .838 & .630 & \underline{.159} & \underline{.654} & .036 & .742 & .822 & \underline{.827} & \underline{.588} \\
  RAFT-collab & .808 & .645 & .147 & .652 & .031 & .708 & \underline{.827} & .796 & .577 \\
  \textbf{C-RL (ours)} & \textbf{.879} & \textbf{.708} & \textbf{.183} & \textbf{.686} & \textbf{.046} & \textbf{.776} & \textbf{.843} & \textbf{.840} & \textbf{.620} \\
\midrule
\multicolumn{10}{@{}l}{\emph{Collaboration protocol: Multiagent Debate}} \\
  Base & .768 & .623 & .155 & .655 & .036 & .720 & .817 & .818 & .574 \\
  GRPO-indiv & .869 & \textbf{.670} & \underline{.177} & \underline{.686} & \underline{.038} & .732 & \underline{.833} & .822 & \underline{.603} \\
  DAPO-indiv & \underline{.879} & .640 & .172 & \textbf{.689} & \underline{.038} & .740 & \underline{.833} & .832 & \underline{.603} \\
  prompt-collab & \underline{.879} & .645 & .139 & .670 & .037 & .744 & .810 & \underline{.838} & .595 \\
  RAFT-collab & .849 & \underline{.655} & .155 & .629 & \underline{.038} & \underline{.748} & .822 & .814 & .589 \\
  \textbf{C-RL (ours)} & \textbf{.899} & \underline{.655} & \textbf{.194} & .664 & \textbf{.044} & \textbf{.764} & \textbf{.845} & \textbf{.846} & \textbf{.614} \\
\midrule
\multicolumn{10}{@{}l}{\emph{Collaboration protocol: SLM-MUX}} \\
  Base & .838 & .568 & .112 & .721 & .079 & .718 & .845 & .856 & .592 \\
  GRPO-indiv & .859 & .588 & \underline{.173} & .721 & .078 & .732 & .845 & .830 & .603 \\
  DAPO-indiv & \underline{.869} & \underline{.650} & .112 & \underline{.750} & .091 & .736 & \underline{.855} & .820 & \underline{.610} \\
  prompt-collab & .828 & .615 & .128 & .714 & \textbf{.109} & \underline{.740} & .850 & \underline{.866} & .606 \\
  RAFT-collab & .838 & .542 & .148 & .661 & .095 & .720 & .848 & .800 & .582 \\
  \textbf{C-RL (ours)} & \textbf{.879} & \textbf{.672} & \textbf{.201} & \textbf{.793} & \underline{.109} & \textbf{.766} & \textbf{.873} & \textbf{.874} & \textbf{.646} \\
\bottomrule
\end{tabular*}
}
\vspace{-5pt}
\caption{\textbf{Collaborative performance}: each policy is evaluated inside the full collaborative system. Best per column within each collaboration protocol in \textbf{bold}, second-best \underline{underlined}; tied values share the same mark. \emph{Avg.} is the mean over all eight datasets. Confidence intervals are reported in Table~\ref{tab:crl-ci}; individual-mode scores in Appendix~\ref{app:crl-indiv}.}
\vspace{-15pt}
\label{tab:crl-collab}
\end{table}

\paragraph{C-RL produces the strongest collaborator under every protocol.} Table~\ref{tab:crl-collab} reports collaborative performance for six methods across three collaboration protocols. C-RL achieves the best score in 21 of the 24 cells, and it is the only method that never degrades collaborative performance relative to the untrained policy: it improves on \emph{Base} in all 24 cells, whereas each of the four alternative baselines regresses somewhere. Averaged over the eight datasets, C-RL adds $+4.8$~pts over \emph{Base} under LLM-Blender, $+4.0$~pts under Multiagent Debate, and $+5.4$~pts under SLM-MUX, and it exceeds the best baseline in each block by $+3.2$~pts, $+1.1$~pts, and $+3.6$~pts respectively. The fact that margins hold across three protocols that compose models in quite different ways indicates that the policy is not exploiting one particular mechanism, but is picking up generalizable collaboration skills and compositional strength through C-RL training. We study generalization to held-out data and collaborations in \S\ref{sec:train-analysis}.

\paragraph{Individual-reward RL can improve a model while making it a worse collaborator.} Optimizing solo correctness does usually help for collaboration: across the 24 cells, GRPO-indiv improves on \emph{Base} in 19 and DAPO-indiv in 21. However, the exceptions show that a policy could get individually \emph{better} and collaboratively \emph{worse} under existing RL pipelines, and the two effects can be badly out of proportion. Together with the individual performance in Appendix~\ref{app:crl-indiv}, DAPO-indiv gains $+6.2$~pts alone on MMLU-ProX yet costs the system $3.6$~pts under SLM-MUX, and GRPO-indiv gains $+2.6$~pts alone on the same dataset while costing the system $2.6$~pts. C-RL never exhibits this: it improves on \emph{Base} in all 24 cells. Optimizing the collaborative outcome directly therefore avoids a failure mode that optimizing solo correctness leaves open. Furthermore, C-RL also improves individual performance in most settings and by a wide margin in places, as a by-product rather than as the objective; Appendix~\ref{app:crl-indiv} reports individual scores showing that it improves on \emph{Base} in 21 of 24 cells, by $+5.3$~pts on average and by as much as $+20.2$~pts on GPQA-Diamond.

\begin{wrapfigure}[27]{r}{0.5\linewidth}
\centering
\vspace{-\baselineskip}
\includegraphics[width=\linewidth]{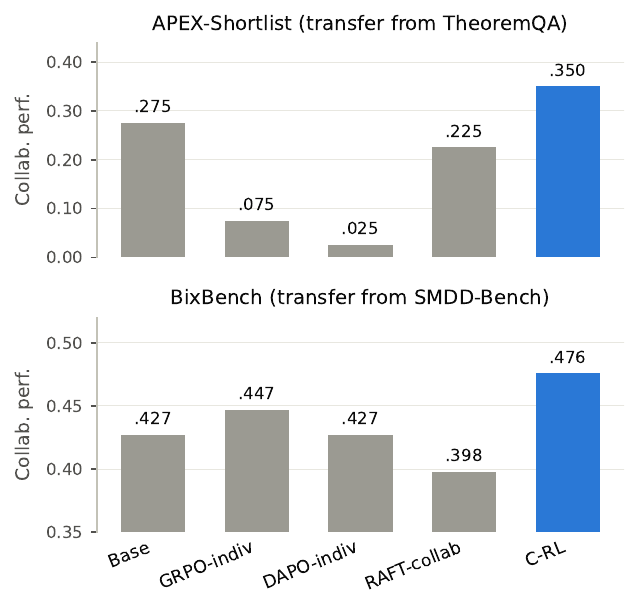}
\caption{Collaboration performance on two held-out datasets, APEX-Shortlist and BixBench. Prompt-collab is not applicable because it performs no training and produces no checkpoint, so it has nothing to transfer.}
\label{fig:heldout-data}
\end{wrapfigure}

\paragraph{The gain concentrates on execution-verified and reasoning tasks.} The size of C-RL's improvement varies considerably by dataset. Relative to \emph{Base} and averaged over the protocols where both are available, the largest relative gain by far is on KernelBench ($+5.6$~pts absolute, $+41\%$ relative), followed by AssayBench ($+1.6$~pts absolute, $+33\%$ relative) and TheoremQA ($+7.7$~pts absolute, $+13\%$ relative). The smallest are MMLU-ProX ($+2.7$~pts absolute, $+3\%$ relative) and CulturalBench-Hard ($+3.1$~pts absolute, $+4\%$ relative). Grouping by how the reward is computed sharpens the pattern: on the three execution-verified and reference-ranking tasks the average relative gain is $+27.0\%$, against $+7.4\%$ on the five answer-matching tasks. Two properties of these tasks plausibly account for this. They ask for open-ended artifacts, a CUDA kernel, a molecule, a ranked gene list, which are scored on a graded scale rather than marked right or wrong, so the reward in \eqref{eq:crl-reward} can register partial improvements that a binary match would discard; and they start from low absolute scores, leaving many problems the system still gets wrong, which is where a stronger policy has the most room to help.

\subsection{Analysis}
\label{sec:train-analysis}

\paragraph{Compositional strength transfers to unseen datasets, while individual-reward RL struggles to.} Section~\ref{sec:train-results} evaluates every policy on the dataset it was trained on, which leaves open whether C-RL teaches the policy something about collaborating or something about the task. To separate the two, we take the checkpoints trained on TheoremQA and SMDD-Bench and evaluate them, with no further training, on two held-out datasets: MathArena APEX-Shortlist \citep{dekoninck2026matharena}, a set of olympiad problems, and BixBench \citep{mitchener2025bixbenchcomprehensivebenchmarkllmbased}, questions in computational biology, using the LLM-Blender collaboration protocol. Figure~\ref{fig:heldout-data} shows that optimizing solo correctness does not merely fail to carry over, it damages the system it is placed in: on APEX-Shortlist, GRPO-indiv falls $20.0$~pts below \emph{Base} and DAPO-indiv $25.0$~pts, and collaborative RAFT falls $5.0$~pts below: these questions are too hard for the policy to tackle individually, and solo RL makes the model focusing more on individually working out problems and failing harder. C-RL is the only method that improves on the untrained policy, by $+7.5$~pts. The ordering repeats on BixBench, where collaborative RAFT again sits below \emph{Base} ($-2.9$~pts), DAPO-indiv exactly matches it, and C-RL leads at $+4.8$~pts. Across both held-out datasets, C-RL is the only method that improves in each. This suggests that C-RL teaches generalizable collaboration skills that carry over to held-out tasks, rather than fitting the policy to the datasets it was trained on.

\begin{wrapfigure}[24]{r}{0.5\linewidth}
\centering
\vspace{-\baselineskip}
\includegraphics[width=\linewidth]{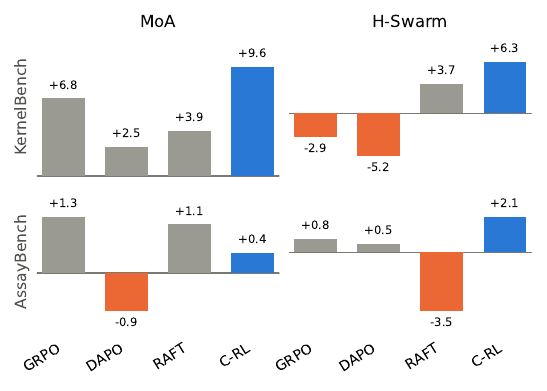}
\vspace*{-20pt}
\caption{Change in collaborative performance relative to \emph{Base} under two protocols never seen during training: Mixture-of-Agents (MoA) and Heterogeneous Swarms (H-Swarm). Bars are GRPO-indiv, DAPO-indiv, RAFT-collab and C-RL, using the KernelBench and AssayBench checkpoints trained under LLM-Blender. Orange marks a method that performs worse than the untrained policy. Prompt-collab is omitted because it trains nothing and has no checkpoint to transfer.}
\label{fig:heldout-collab}
\end{wrapfigure}

\paragraph{With C-RL, compositional strength also transfers to unseen collaboration protocols.} The same question can be asked of the collaboration protocol: does a policy trained inside one protocol remain helpful in another collaboration method? We take the KernelBench and AssayBench checkpoints trained under LLM-Blender and evaluate them, without further training, under two protocols held out from \S\ref{sec:train-setup}: Mixture-of-Agents \citep{wang2025mixture}, which layers several rounds of proposers over an aggregator, and Heterogeneous Swarms \citep{feng2025heterogeneous}, which searches over roles and weights in a graph of models. Figure~\ref{fig:heldout-collab} reports the result. C-RL is the best method in three of the four cells, and it is again the only method that improves on \emph{Base} everywhere: it adds $+9.6$~pts on KernelBench under Mixture-of-Agents, $+6.3$~pts on KernelBench under Heterogeneous Swarms, and $+2.1$~pts on AssayBench under Heterogeneous Swarms. The baselines are again erratic across the switch. DAPO-indiv falls $5.2$~pts below \emph{Base} on KernelBench under Heterogeneous Swarms and GRPO-indiv $2.9$~pts below, and collaborative RAFT, the strongest baseline on that same cell, collapses to $0.011$ on AssayBench under the same protocol, $3.5$~pts below \emph{Base}. The one cell C-RL does not win is AssayBench under Mixture-of-Agents, where GRPO-indiv reaches $0.076$ against C-RL's $0.068$, still close. Taken with Figure~\ref{fig:heldout-data}, the policy carries its helpfulness in collaboration across both axes, the dataset and the protocol, indicating that C-RL teaches generalizable collaboration skills rather than overfitting to the training environment.

\begin{wraptable}[17]{r}{0.42\linewidth}
\centering
\vspace{-\baselineskip}
\footnotesize
\setlength{\tabcolsep}{3pt}
\begin{tabular}{@{}l cc @{\hspace{7pt}} cc@{}}
\toprule
\multirow{2}{*}{Method} & \multicolumn{2}{c}{KernelBench} & \multicolumn{2}{c}{AssayBench} \\
\cmidrule(lr){2-3} \cmidrule(lr){4-5}
 & score & $\Delta$ & score & $\Delta$ \\
\midrule
  Base & .174 & -- & .044 & -- \\
  GRPO-indiv & .185 & $+1.1$ & .033 & $-1.1$ \\
  DAPO-indiv & .223 & $+4.9$ & .026 & $-1.9$ \\
  prompt-collab & .171 & $-0.3$ & .043 & $-0.2$ \\
  RAFT-collab & .215 & $+4.1$ & .032 & $-1.2$ \\
\midrule
  \textbf{C-RL (ours)} & \textbf{.254} & $+8.0$ & \textbf{.055} & $+1.1$ \\
\bottomrule
\end{tabular}
\caption{\textbf{Frontier collaborators.} Collaborative performance with GPT-5.5, Gemini 3.5 Flash and Claude Opus 4.8 as the pool, under LLM-Blender. $\Delta$ is the change in percentage points relative to the untrained \emph{Base} policy in the same pool. Best per dataset in \textbf{bold}.}
\vspace{-10pt}
\label{tab:crl-sota}
\end{wraptable}

\paragraph{C-RL empowers small policy to collaborate with state-of-the-art proprietary LLMs.} We replace the three collaborators with frontier proprietary systems, GPT-5.5, Gemini 3.5 Flash and Claude Opus 4.8, retrain under LLM-Blender, and evaluate on KernelBench and AssayBench. The stronger collaborator pool does raise the ceiling: the untrained policy's system scores $0.174$ and $0.044$, against $0.144$ and $0.035$ with the original open-model collaborators. Table~\ref{tab:crl-sota} shows that C-RL still adds to it, and by more than any baseline, reaching $0.254$ on KernelBench ($+8.0$~pts over \emph{Base}) and $0.055$ on AssayBench ($+1.1$~pts). AssayBench separates the methods most sharply: every baseline leaves the system worse than the untrained policy does, by $0.2$ to $1.9$~pts, so solo RL actually makes collaboration with state-of-the-art models worst on this dataset. Only C-RL improves on it: a policy small enough to be trained cheaply can therefore be useful to a system built from models far larger than itself, but only if it is trained for what it contributes rather than for what it can do alone.

\section{Related Work}
\label{sec:related}

\paragraph{Multi-LLM collaboration.} A growing body of work builds systems in which several language models solve a task together rather than one model solving it alone \citep{du2024position, feng2026one}, a shift from models to compound systems \citep{zaharia2024compound}. Approaches differ in what the models exchange. Routing and cascading methods dispatch each query to a single model, selecting by learned preference \citep{ong2025routellm}, graph structure \citep{fenggraphrouter}, generation-time switching \citep{feng2025don}, prompted choice \citep{frick2025prompt}, cost-aware cascades \citep{chen2024frugalgpt, dingbest}, or inferred skill requirements \citep{chen2025symbolic}. Agentic methods let models exchange text over several rounds, through debate \citep{du2023improving}, feedback \citep{feng2024don}, optimizable communication graphs \citep{zhugegptswarm}, recruited expert teams \citep{chen2024agentverse}, sequential context passing \citep{zhang2024chain}, or swarm search over roles \citep{feng2025heterogeneous}. A third family combines outputs or parameters directly, by ranking and fusing candidates \citep{jiang2023llm}, layering mixtures of agents \citep{wang2025mixture}, training aggregators \citep{zhao2025majority}, fine-tuning within a society of models \citep{subramaniammultiagent}, or aligning a pool collectively \citep{jiang2025sparta}. Two distinctions separate our setting from much of this work. First, a large part of the multi-agent literature instantiates a \emph{single} model repeatedly under different prompts, personas, or manually crafted roles, so the participants differ in their instructions rather than in their parameters. We study pools of genuinely distinct models, which is what makes contribution a property of a model rather than of a prompt. Second, alongside methods that compose frozen models at inference time, we also include collaboration methods that update model weights, i.e., training-time model collaboration, so the pool is not required to stay fixed. Across this entire literature, it is the collaboration strategy that gets optimized: making models better participants and collaborators, independent of which strategy it is placed into, is a separate research question that we address with C-RL.

\paragraph{Evaluating models alone and in systems.} Language models are predominantly evaluated on their own, against benchmarks and leaderboards \citep{liang2023holistic, chiang2024chatbot}. Another line evaluates models as parts of systems, finding that measured capability depends on the interactive setting \citep{lee2022evaluating}, the agent scaffold \citep{kapoor2024ai, kapoor2026holistic}, and the surrounding harness \citep{yao2026harness}. A third asks when collaboration is worthwhile: \citet{pappu2026multiagent} report cases where a team falls short of its strongest member, \citet{cemri2026multi} catalogue how multi-agent systems fail, \citet{kim2026capable} study how the benefit of coordination varies with the capability of the models involved, and \citet{zhang2026mixture} argue that a model's value lies in its complementarity with others. \citet{sanchez2025llm} likewise estimate synergistic and antagonistic interaction between models in order to recommend which ensemble to deploy. While previous work focuses on one ensemble or task at a time, we make it general: we measure a model's contribution across many pools, tasks, and collaboration methods, and ask directly whether it can be predicted from the model's own benchmark score.

\paragraph{Attributing contribution.} Dividing a collective outcome among participants is the classical problem addressed by the Shapley value \citep{shapley1953value} and the Banzhaf value \citep{banzhaf1964weighted}, both of which are widely used in machine learning \citep{lundberg2017unified, ijcai2022p778}. Most applications value training data \citep{ghorbani2019data, pmlr-v151-kwon22a, wang2023data}, where exact computation is intractable and much of the effort goes into cheaper estimators \citep{wang2024helpful, wang2025data}. The relation between leave-one-out marginals and full Shapley values has been studied directly: \citet{pmlr-v151-kwon22a} place both within a single semivalue family, \citet{wang2024helpful} find leave-one-out less robust than Shapley for attributing individual training instances, and \citet{lu2026agents} find it sufficient for identifying weak agents in multi-agent LLM systems. Within language-model systems, attribution has been applied to prompt tokens \citep{horovicz2024tokenshap}, workflow modules \citep{yang2025understanding}, and agents \citep{lu2026agents, hua2026shapley, zhang2025agent}, in each case to diagnose or improve one particular system. We instead treat a model's marginal contribution as a property of the model, measure it across many pools, tasks, and collaboration methods, and then optimize it directly.

\paragraph{Reinforcement learning for language models.} Post-training with RL is standard practice, from preference feedback \citep{ouyang2022training, rafailov2023direct} to verifiable rewards \citep{guo2025deepseek, lambert2024tulu}, commonly optimized with group-relative estimators \citep{shao2024deepseekmath, yu2026dapo} built on clipped policy gradients \citep{schulman2017proximal}. Extending these methods to multi-turn and agentic settings raises questions of credit assignment \citep{zhou2024archer, feng2026group}, including whether group-relative baselines remain appropriate when prompts differ across roles and turns \citep{zhao2025stronger}. Several lines of work apply reinforcement learning to multi-model systems. Some co-train every participant against the collaborative outcome \citep{park2025maporl, liu2026llm, estornell2025acc, motwani2024malt}; some train a model against copies of itself in self-play \citep{liu2026spiral, ma2024coevolving, zhao2026absolute}; some distill a multi-agent system into a single model \citep{li2025chain}; and some train only the component that routes or aggregates while the rest of the pool stays frozen \citep{estornell2025train, zhang2026router, fu2025rlae}. Unlike any of these, our collaborators are neither trained nor copies of the policy, and the per-sample reward is the system's raw outcome itself, with the policy's marginal contribution recovered only at the group-relative advantage, never as a per-sample reward transformation. As a result, C-RL can be flexibly applied to diverse model/task/protocol combinations, optimize compositional strength directly, and enhance the policy as better collaborators in general. The name \emph{compositional reinforcement learning} has separately been used for composing training prompts into harder ones \citep{xu2026composition} and for decomposing a single-agent control task via logical specifications \citep{jothimurugan2021compositional}; both target a single policy acting alone, whereas our C-RL trains a policy against the outcome of a live multi-model collaboration.

\section{Conclusion}
\label{sec:conclusion}

We propose to evaluate and train language models by their \emph{compositional strength}, how much a model adds to the multi-model system it belongs to rather than how well it performs alone. We formalize it as a model's Shapley value in a cooperative game over coalitions of a model pool, and estimate it with a leave-one-out marginal. Extensive experiments show that compositional strength is largely uncorrelated with individual strength, so the benchmarks and leaderboards that guide model selection say little about how helpful they are in collaborative systems. We therefore optimize it directly with compositional reinforcement learning (C-RL), which rewards a policy with the outcome of the system it participates in. C-RL produces the strongest collaborator in 21 of 24 settings, never leaves a system worse off than the untrained policy, and carries over to held-out datasets, unseen protocols, and pools of far stronger proprietary models. We see compositional strength as a step toward system-level intelligence, where models are built, evaluated, and trained not for what they can do alone, but for what they make possible together.

\subsection*{AI use statement}

We used generative AI tools to help with language polishing. The AI tools were used solely for language editing and did not contribute to the intellectual content of the work. We have reviewed all AI-assisted edits and take responsibility for the final content of this work.

\subsection*{Reproducibility statement}

Model pools, collaboration methods, and datasets for the evaluation study are listed in \S\ref{sec:eval-setup}, with dataset splits and pool composition in Appendix~\ref{app:eval-setup}; the policy, collaborators, protocols, tasks, and baselines for the training study are listed in \S\ref{sec:train-setup}, with hyperparameters, implementation, and checkpoint selection in Appendix~\ref{app:crl-impl}. Confidence intervals are in Appendix~\ref{app:crl-ci}. One caveat limits exact reproduction: several collaborators are proprietary models served behind APIs whose updates we do not control, so numbers involving them may shift as APIs change. We will release code for both the evaluation and training studies, together with the configurations needed to reproduce every reported run, upon acceptance.

\subsection*{Ethics Statement}

C-RL's objective, rewarding a policy for its contribution to a collaborative system's outcome, is dual-use: the same training signal could let a malicious model more effectively inject harmful intent into a multi-model system while remaining individually inconspicuous, compounding the risks studied in adversarial collaboration settings \citep{yang2026among}. Enhancing the safety of collaborative systems against such malicious contributions is an important direction for future work.

\bibliography{iclr2027_conference}
\bibliographystyle{iclr2027_conference}

\appendix
\section{Additional Results and Analysis}
\label{app:results}

\subsection{Individual Performance}
\label{app:crl-indiv}

\begin{table}[t]
\centering
\footnotesize
\setlength{\tabcolsep}{0.6pt}
\begin{tabular}{@{}l cccccccc @{\hspace{6pt}}c@{}}
\toprule
Method & GPQA-dia & Theorem & KernelB & SMDD & AssayB & ChemB & Culture-H & MMLU-PX & Avg. \\
\midrule
  Base & .687 & .547 & .103 & .279 & .023 & .694 & .783 & .758 & .484 \\
  GRPO-indiv & .808 & .608 & .111 & .322 & .026 & .714 & .832 & .784 & .526 \\
  DAPO-indiv & .758 & .640 & .114 & .261 & .024 & .724 & .819 & .820 & .520 \\
  prompt-collab & .697 & .580 & .110 & .260 & .026 & .700 & .818 & .796 & .498 \\
  RAFT-collab & .768 & .637 & .108 & \textbf{.410} & .017 & .684 & .806 & .764 & .524 \\
\midrule
  \textbf{C-RL}, LLM-Blender & \textbf{.889} & \textbf{.690} & .097 & .297 & .024 & .710 & .842 & \textbf{.826} & \textbf{.547} \\
  \textbf{C-RL}, MA Debate & .818 & .640 & \textbf{.115} & .340 & \textbf{.029} & .750 & .834 & \textbf{.826} & .544 \\
  \textbf{C-RL}, SLM-MUX & .748 & .552 & .094 & .358 & .023 & \textbf{.752} & \textbf{.843} & .810 & .523 \\
\bottomrule
\end{tabular}
\caption{\textbf{Individual performance}: the trained policy answers alone, outside any collaborative system. The five baselines do not depend on the collaboration protocol and are reported once; C-RL is reported per training protocol. Best per column in \textbf{bold}. \emph{Avg.} is the mean over all eight datasets.}
\label{tab:crl-indiv}
\end{table}

Table~\ref{tab:crl-indiv} scores every method with the policy answering alone, outside any collaborative system. Because a policy's solo score does not depend on the protocol it was later evaluated in, the five baselines are reported once; only C-RL is trained separately per protocol.

Read against Table~\ref{tab:crl-collab}, these numbers make the decoupling of \S\ref{sec:eval} visible within the training experiment rather than across a pool of frozen models. Ranking the five baselines by individual performance does not reproduce their ranking as collaborators. Collaborative RAFT is among the strongest baselines in individual mode (average $0.524$, essentially tied with GRPO-indiv at $0.526$), yet it is the weakest collaborator overall, improving on \emph{Base} in only 14 of 24 cells and finishing last or second to last under all three protocols. Prompt nudge is the mirror image: it changes no weights and sits near the bottom individually ($0.498$, against $0.484$ for \emph{Base}), yet it is the best baseline under LLM-Blender and the second best under SLM-MUX. The same gap holds for C-RL itself, which leads in individual mode in 13 of the 24 cells against 21 of 24 in collaborative mode. A method's individual standing is thus a poor guide to its collaborative standing in either direction, which is why \S\ref{sec:train-results} reads C-RL's individual gains as a by-product rather than as the mechanism behind its collaborative ones.

\subsection{Confidence Intervals}
\label{app:crl-ci}

Table~\ref{tab:crl-ci} places the untrained policy beside C-RL under each protocol, with a $95\%$ confidence interval on every C-RL score. The intervals are computed at each task's true test-set size, which ranges from $99$ items on GPQA-Diamond to $613$ on CulturalBench-Hard, so they capture how much of a gap is attributable to having scored a finite test set. A star marks the cells where the lower end of the interval still sits above the \emph{Base} score, which holds for the aggregate column under all three protocols; the exceptions among individual datasets are concentrated on the two smallest and noisiest tasks. We report these for transparency about precision rather than as hypothesis tests: the claim in \S\ref{sec:train-results} rests on C-RL winning $21$ of $24$ cells and never falling below \emph{Base}, a pattern across cells that no single interval speaks to.

\begin{table}[t]
\centering
\footnotesize
\setlength{\tabcolsep}{0pt}
\begin{tabular*}{\linewidth}{@{\extracolsep{\fill}}l cccccccc c@{}}
\toprule
Method & GPQA-dia & Theorem & KernelB & SMDD & AssayB & ChemB & Culture-H & MMLU-PX & Avg. \\
\midrule
\multicolumn{10}{@{}l}{\emph{Collaboration protocol: LLM-Blender}} \\
  Base & .818 & .613 & .144 & .635 & .035 & .720 & .806 & .804 & .572 \\
  \multirow{2}{*}{\textbf{C-RL (ours)}} & \textbf{.879} & \textbf{.708}$^{*}$ & \textbf{.183} & \textbf{.686} & \textbf{.046} & \textbf{.776}$^{*}$ & \textbf{.843}$^{*}$ & \textbf{.840}$^{*}$ & \textbf{.620}$^{*}$ \\
   & {\tiny[.808,.939]} & {\tiny[.662,.752]} & {\tiny[.120,.256]} & {\tiny[.606,.759]} & {\tiny[.024,.069]} & {\tiny[.738,.812]} & {\tiny[.814,.873]} & {\tiny[.808,.872]} & {\tiny[.602,.638]} \\[2pt]
\midrule
\multicolumn{10}{@{}l}{\emph{Collaboration protocol: Multiagent Debate}} \\
  Base & .768 & .623 & .155 & .655 & .036 & .720 & .817 & .818 & .574 \\
  \multirow{2}{*}{\textbf{C-RL (ours)}} & \textbf{.899}$^{*}$ & \textbf{.655} & \textbf{.194} & \textbf{.664} & \textbf{.044} & \textbf{.764}$^{*}$ & \textbf{.845} & \textbf{.846} & \textbf{.614}$^{*}$ \\
   & {\tiny[.838,.949]} & {\tiny[.608,.703]} & {\tiny[.128,.264]} & {\tiny[.584,.745]} & {\tiny[.024,.069]} & {\tiny[.726,.800]} & {\tiny[.816,.873]} & {\tiny[.814,.876]} & {\tiny[.596,.632]} \\[2pt]
\midrule
\multicolumn{10}{@{}l}{\emph{Collaboration protocol: SLM-MUX}} \\
  Base & .838 & .568 & .112 & .721 & .079 & .718 & .845 & .856 & .592 \\
  \multirow{2}{*}{\textbf{C-RL (ours)}} & \textbf{.879} & \textbf{.672}$^{*}$ & \textbf{.201}$^{*}$ & \textbf{.793}$^{*}$ & \textbf{.109} & \textbf{.766}$^{*}$ & \textbf{.873}$^{*}$ & \textbf{.874} & \textbf{.646}$^{*}$ \\
   & {\tiny[.808,.939]} & {\tiny[.625,.718]} & {\tiny[.136,.272]} & {\tiny[.723,.861]} & {\tiny[.078,.144]} & {\tiny[.728,.802]} & {\tiny[.847,.899]} & {\tiny[.844,.902]} & {\tiny[.628,.663]} \\[2pt]
\bottomrule
\end{tabular*}
\caption{\textbf{Confidence intervals for C-RL.} For each collaboration protocol we report the untrained \emph{Base} policy alongside \emph{C-RL}, with 95\% confidence intervals for C-RL in brackets. $^{*}$ marks cells where the lower bound of the C-RL interval still exceeds the Base score.}
\label{tab:crl-ci}
\end{table}

\subsection{Response-Level Analysis}
\label{app:response-analysis}

The results in \S\ref{sec:train-results} are aggregate scores. To see what changes inside a rollout, we examine the per-item logs of every (protocol, task) pair for the untrained policy and for C-RL.

\begin{table}[t]
\centering
\footnotesize
\setlength{\tabcolsep}{2.5pt}
\begin{tabular*}{\linewidth}{@{\extracolsep{\fill}}l cc cc cc@{}}
\toprule
& \multicolumn{2}{c}{LLM-Blender} & \multicolumn{2}{c}{Multiagent Debate} & \multicolumn{2}{c}{SLM-MUX} \\
\cmidrule(lr){2-3} \cmidrule(lr){4-5} \cmidrule(lr){6-7}
Task & rescue & drag & rescue & drag & rescue & drag \\
\midrule
  GPQA-dia & .162$\,$\textcolor{gray}{$\to$}$\,$.040 & .030$\,$\textcolor{gray}{$\to$}$\,$.051 & .152$\,$\textcolor{gray}{$\to$}$\,$.111 & .071$\,$\textcolor{gray}{$\to$}$\,$.030 & .172$\,$\textcolor{gray}{$\to$}$\,$.172 & .020$\,$\textcolor{gray}{$\to$}$\,$.040 \\
  Theorem & .098$\,$\textcolor{gray}{$\to$}$\,$.048 & .033$\,$\textcolor{gray}{$\to$}$\,$.030 & .100$\,$\textcolor{gray}{$\to$}$\,$.052 & .025$\,$\textcolor{gray}{$\to$}$\,$.037 & .085$\,$\textcolor{gray}{$\to$}$\,$.152 & .065$\,$\textcolor{gray}{$\to$}$\,$.033 \\
  KernelB & .136$\,$\textcolor{gray}{$\to$}$\,$.168 & .064$\,$\textcolor{gray}{$\to$}$\,$.032 & .144$\,$\textcolor{gray}{$\to$}$\,$.160 & .056$\,$\textcolor{gray}{$\to$}$\,$.024 & .140$\,$\textcolor{gray}{$\to$}$\,$.220 & .120$\,$\textcolor{gray}{$\to$}$\,$.020 \\
  SMDD & .438$\,$\textcolor{gray}{$\to$}$\,$.438 & .080$\,$\textcolor{gray}{$\to$}$\,$.088 & .474$\,$\textcolor{gray}{$\to$}$\,$.423 & .095$\,$\textcolor{gray}{$\to$}$\,$.095 & .504$\,$\textcolor{gray}{$\to$}$\,$.526 & .051$\,$\textcolor{gray}{$\to$}$\,$.073 \\
  AssayB & .006$\,$\textcolor{gray}{$\to$}$\,$.009 & .000$\,$\textcolor{gray}{$\to$}$\,$.000 & .000$\,$\textcolor{gray}{$\to$}$\,$.000 & .000$\,$\textcolor{gray}{$\to$}$\,$.003 & -- & -- \\
  ChemB & .098$\,$\textcolor{gray}{$\to$}$\,$.120 & .072$\,$\textcolor{gray}{$\to$}$\,$.054 & .104$\,$\textcolor{gray}{$\to$}$\,$.064 & .078$\,$\textcolor{gray}{$\to$}$\,$.050 & .112$\,$\textcolor{gray}{$\to$}$\,$.058 & .038$\,$\textcolor{gray}{$\to$}$\,$.044 \\
  Culture-H & .077$\,$\textcolor{gray}{$\to$}$\,$.034 & .054$\,$\textcolor{gray}{$\to$}$\,$.033 & .086$\,$\textcolor{gray}{$\to$}$\,$.034 & .052$\,$\textcolor{gray}{$\to$}$\,$.023 & .096$\,$\textcolor{gray}{$\to$}$\,$.054 & .034$\,$\textcolor{gray}{$\to$}$\,$.010 \\
  MMLU-PX & .086$\,$\textcolor{gray}{$\to$}$\,$.048 & .040$\,$\textcolor{gray}{$\to$}$\,$.034 & .092$\,$\textcolor{gray}{$\to$}$\,$.044 & .032$\,$\textcolor{gray}{$\to$}$\,$.024 & .112$\,$\textcolor{gray}{$\to$}$\,$.102 & .014$\,$\textcolor{gray}{$\to$}$\,$.038 \\
\midrule
  \emph{Mean} & .137$\,$\textcolor{gray}{$\to$}$\,$.113 & .047$\,$\textcolor{gray}{$\to$}$\,$.040 & .144$\,$\textcolor{gray}{$\to$}$\,$.111 & .051$\,$\textcolor{gray}{$\to$}$\,$.036 & .174$\,$\textcolor{gray}{$\to$}$\,$.183 & .049$\,$\textcolor{gray}{$\to$}$\,$.037 \\
\bottomrule
\end{tabular*}
\caption{\textbf{Where the policy and the system disagree.} Each cell reads \emph{base}$\to$\emph{C-RL}. \emph{Rescue} is the fraction of test items the policy answers incorrectly on its own but the system gets right; \emph{drag} is the fraction the policy answers correctly on its own but the system gets wrong. Cells marked \texttt{--} lack a usable log. Drag falls under all three protocols.}
\label{tab:rescue-drag}
\end{table}

\paragraph{C-RL makes the policy's correct answers more likely to survive.} Pairing each item's individual-mode score with its collaborative-mode score partitions the test set by where the policy and the system disagree. Two cells are informative. \emph{Rescue} is the fraction of items the policy gets wrong alone but the system gets right, the case \S\ref{sec:train-method} appeals to when it argues that a policy can be rewarded without being individually correct. \emph{Drag} is the opposite, items the policy gets right alone but the system gets wrong, which is information the collaboration destroys. Table~\ref{tab:rescue-drag} reports both. Drag falls under all three protocols, from $4.7$ to $4.0$ percent of items under LLM-Blender, $5.1$ to $3.6$ under Multiagent Debate, and $4.9$ to $3.7$ under SLM-MUX. Rescue stays substantial throughout, between $11$ and $18$ percent of items, so the policy continues to be carried by its collaborators on a large share of questions even after training; what changes is that it is dragged down less often. Rescue itself falls slightly under LLM-Blender and Multiagent Debate, which is expected rather than informative, since a policy that answers more questions correctly on its own leaves fewer left to be rescued.

\begin{table}[t]
\centering
\footnotesize
\setlength{\tabcolsep}{2pt}
\begin{tabular*}{\linewidth}{@{\extracolsep{\fill}}l cccc c c@{}}
\toprule
& \multicolumn{4}{c}{option proposed by each model} & \multirow{2}{*}{system} & \multirow{2}{*}{policy len.} \\
\cmidrule(lr){2-5}
& Policy & gpt-oss & kimi-k3 & nemotron & & \\
\midrule
\multicolumn{7}{@{}p{0.97\linewidth}@{}}{What happens if aromatic nitro compounds are mixed with sodium hydroxide? \textbf{(A)}~Nothing happens \textbf{(B)}~They may explode \textbf{(C)}~They decompose \textbf{(D)}~They form a salt~ Correct: \textbf{B}} \\[2pt]
  Base & A & A & \textbf{B} & A & A & 34,350 \\
  \textbf{C-RL} & \textbf{B} & A & \textbf{B} & A & \textbf{B} & 4,059 \\
\midrule
\multicolumn{7}{@{}p{0.97\linewidth}@{}}{Which of the following statements is true about halogenated organic compounds? \textbf{(A)}~They are incompatible with strong oxidizing and reducing agents. \textbf{(B)}~They react slowly with metals. \textbf{(C)}~The more halogen atoms, the more reactive they become. \textbf{(D)}~They are highly reactive.~ Correct: \textbf{A}} \\[2pt]
  Base & D & \textbf{A} & \textbf{A} & \textbf{A} & D & 7,292 \\
  \textbf{C-RL} & \textbf{A} & \textbf{A} & \textbf{A} & \textbf{A} & \textbf{A} & 6,811 \\
\bottomrule
\end{tabular*}
\caption{\textbf{Two ChemBench items where C-RL changes the system's answer.} Each model proposes a candidate and a fuser combines them; the letter shown is the option each candidate ends on, with the correct option in \textbf{bold}. In the first item the untrained policy proposes the wrong option and the system follows it; after C-RL the policy proposes the correct option and the system does too. In the second the untrained policy dissents from three collaborators that are all correct and the system adopts the policy's error. The last column gives the length of the policy's own candidate.}
\label{tab:qualitative}
\end{table}

\paragraph{Two illustrative items.} Table~\ref{tab:qualitative} shows two ChemBench questions under LLM-Blender where C-RL changes the system's answer, with each model's proposed option and the option the fuser settles on. In the first, the untrained policy proposes the wrong option and only one collaborator proposes the right one, and the system follows the majority into the wrong answer; after C-RL the policy proposes the correct option, and the system follows. In the second, the untrained policy dissents from three collaborators that all propose the correct option, and the system nonetheless adopts the policy's error, an instance of drag in which one participant overturns a system that would otherwise have been right. After C-RL the policy agrees and the answer stands. The two cases are the mechanism behind the aggregate numbers: the policy's contribution is not merely more often correct, it is weighted differently by the protocol that consumes it. The length column also records a side effect visible throughout the logs. The untrained policy's candidate in the first item runs to $34{,}350$ characters, against $4{,}059$ after C-RL, so training against the system's outcome shortens contributions that the fuser has to read.

\begin{table}[t]
\centering
\footnotesize
\begin{tabular}{@{}p{0.06\linewidth}p{0.90\linewidth}@{}}
\toprule
\multicolumn{2}{@{}p{0.96\linewidth}@{}}{\textbf{Q.} How many ways are there to arrange the letters in the word *BANANA* up to the symmetries of the word?} \\[3pt]
  Base & The phrase "up to the symmetries" implies treating the arrangement as a bracelet, where two arrangements are considered identical if one can be rotated or flipped to match the other (the dihedral group $D_6$). Applying Burnside's Lemma, the identity symmetry fixes all 60 distinct linear permutations, while the five non-trivial rotations fix none because the letter counts (3 A's, 2 N's, 1 B) cannot form the required periodic patterns. Among the six reflections, the three axes passing through opposite letter positions each fix 4 arrangements (by assigning the AA and NN pairs to the two swapped slots and the remaining A and B to the fixed slots), while the three axes passing between slots fix none. Summing the fixed counts gives $60 + 3 \times 4 = 72$, and dividing by the 12 total symmetries yields 6 distinct arrangements. Final Answer: \boxed{6} \\[3pt]
  \textbf{C-RL} & First, calculate the number of distinct linear permutations of the letters in BANANA, which is $6!/(3!\,2!) = 60$. The phrase "up to symmetries" for a linear word means we consider an arrangement equivalent to its reverse. Since the letter counts for 'A' (3) and 'B' (1) are odd, no arrangement can be a palindrome, meaning every permutation is distinct from its reverse and forms a unique pair. Dividing the total linear permutations by 2 to account for this reversal symmetry gives 30 distinct arrangements. Final Answer: \boxed{30} \\[3pt]
\midrule
\multicolumn{2}{@{}p{0.96\linewidth}@{}}{\textbf{Q.} Determine the product obtained when 3,4-dimethylhexanedial is reacted with the following set of reagents: 1. KOH, H2O, THF, Heat 2. CH3CH2MgBr, H3O+ 3. PCC, CH2Cl2 4. O3, H2O \newline \textbf{A.}~3,4-dimethyl-5,6-dioxooctanoic acid \quad $\ldots$ \quad \textbf{D.}~3,4-dimethyl-5,6-dioxooctanal} \\[3pt]
  Base & Base-catalyzed aldol condensation of 3,4-dimethylhexanedial forms a cyclohexene derivative, which subsequently undergoes Grignard addition and PCC oxidation to introduce an ethyl ketone group. Ozonolysis cleaves the ring double bond to yield an acyclic chain containing aldehyde and ketone functionalities. The resulting 8-carbon skeleton and carbonyl positions precisely match 3,4-dimethyl-5,6-dioxooctanal, while ozonolysis of an internal alkene does not produce carboxylic acids, eliminating option A. Answer: D \\[3pt]
  \textbf{C-RL} & 3,4-Dimethylhexanedial undergoes an intramolecular aldol condensation under basic conditions to form a five-membered cyclic $\alpha$,$\beta$-unsaturated aldehyde. Subsequent Grignard addition and PCC oxidation convert the side-chain aldehyde into an ethyl ketone. Finally, oxidative ozonolysis cleaves the ring double bond, opening the ring to yield a linear dicarbonyl carboxylic acid chain. Counting the carbons and functional groups confirms the product is 3,4-dimethyl-5,6-dioxooctanoic acid. Answer: A \\[3pt]
\bottomrule
\end{tabular}
\caption{\textbf{Collaborative responses the system produces, before and after C-RL.} Two items the untrained system answers incorrectly and the C-RL system answers correctly, quoted in full from the logs (the fuser's internal deliberation is omitted). Both are under LLM-Blender, on TheoremQA and GPQA-Diamond respectively; in both the disagreement is interpretive rather than arithmetic. The second item's two remaining distractors are elided: they are identical to one another in the source dataset and neither system proposes either of them.}
\label{tab:examples}
\end{table}

\paragraph{What the system's answers look like.} Table~\ref{tab:examples} quotes two collaborative responses in full, on items the untrained system gets wrong and the C-RL system gets right. Neither difference is a matter of arithmetic. In the TheoremQA item, the two systems disagree about what the question asks: the untrained system reads ``up to the symmetries of the word'' as the symmetries of a bracelet, invokes the dihedral group and Burnside's lemma, and answers $6$, while the C-RL system reads the word as a linear string with only reversal symmetry, notes that no arrangement of BANANA is a palindrome, and divides the $60$ linear permutations by two to reach $30$. In the GPQA item the two agree on the whole reaction sequence and part company at the last step, where the untrained system asserts that ozonolysis ``does not produce carboxylic acids'' and answers with the aldehyde, while the C-RL system recognises that the aqueous workup specified in the question makes the ozonolysis oxidative and answers with the acid.

We read this as evidence that what C-RL changes is not a single skill but the way the policy's contribution is framed for the models that consume it. Both systems have the same collaborators and the same protocol; what differs is the candidate the policy supplies, and in both cases the decisive content is an interpretive commitment made early, about what is being asked or which variant of a reaction applies, that the rest of the system then follows. This is also why compositional strength is not a restatement of individual strength: the policy's influence runs through whether its framing is adopted by the mechanism that reads it, and Table~\ref{tab:rescue-drag} shows that an untrained policy which is individually correct can still push a system to the wrong answer.

\subsection{Collaboration Effectiveness by Pool and Domain}
\label{app:collab-effective}

Table~\ref{tab:collab-effective} lists the 25 (model pool, task category) values underlying the collaboration-effectiveness claim in \S\ref{sec:eval-results}: for each cell, the leave-one-out and full-coalition scores and the resulting gain, for the (dataset, method) pair that maximizes it.

\begin{table}[t]
\centering
\footnotesize
\setlength{\tabcolsep}{5pt}
\begin{tabular}{@{}lccccc@{}}
\toprule
Model pool & QA & Reasoning & Coding & Pluralism \& Safety & Science \\
\midrule
Specialized Expert & \shortstack{65.8$\to$72.4\\(+6.6)} & \shortstack{32.3$\to$45.5\\(+13.1)} & \shortstack{44.1$\to$65.5\\(+21.4)} & \shortstack{38.7$\to$54.0\\(+15.3)} & \shortstack{41.2$\to$55.4\\(+14.2)} \\[6pt]
8-30B Open & \shortstack{82.9$\to$94.9\\(+12.0)} & \shortstack{24.2$\to$35.6\\(+11.4)} & \shortstack{70.7$\to$82.5\\(+11.8)} & \shortstack{78.2$\to$81.2\\(+3.1)} & \shortstack{57.7$\to$60.8\\(+3.1)} \\[6pt]
$>$100B Open & \shortstack{80.0$\to$83.7\\(+3.7)} & \shortstack{48.1$\to$60.0\\(+11.9)} & \shortstack{48.8$\to$71.2\\(+22.4)} & \shortstack{77.0$\to$82.0\\(+4.9)} & \shortstack{53.7$\to$68.3\\(+14.6)} \\[6pt]
Proprietary API & \shortstack{86.8$\to$90.0\\(+3.2)} & \shortstack{58.5$\to$69.4\\(+10.9)} & \shortstack{66.7$\to$94.1\\(+27.5)} & \shortstack{32.1$\to$37.7\\(+5.6)} & \shortstack{73.1$\to$80.6\\(+7.5)} \\[6pt]
Diverse Academic & \shortstack{66.1$\to$70.0\\(+3.9)} & \shortstack{32.4$\to$65.3\\(+32.9)} & \shortstack{37.4$\to$52.7\\(+15.4)} & \shortstack{64.6$\to$71.6\\(+7.0)} & \shortstack{50.4$\to$55.1\\(+4.7)} \\
\bottomrule
\end{tabular}
\caption{The best-performing collaboration method across 25 (model pool, task category) values underlying the collaboration-effectiveness claim in \S\ref{sec:eval-results}. The mean of the 25 gains is $+11.5$ with a $95\%$ CI of $[+8.7, +14.7]$.}
\label{tab:collab-effective}
\end{table}

\subsection{Per-Dataset Gain Consistency}
\label{app:gain-consistency}

\begin{figure}[h]
\begin{center}
\includegraphics[width=\linewidth]{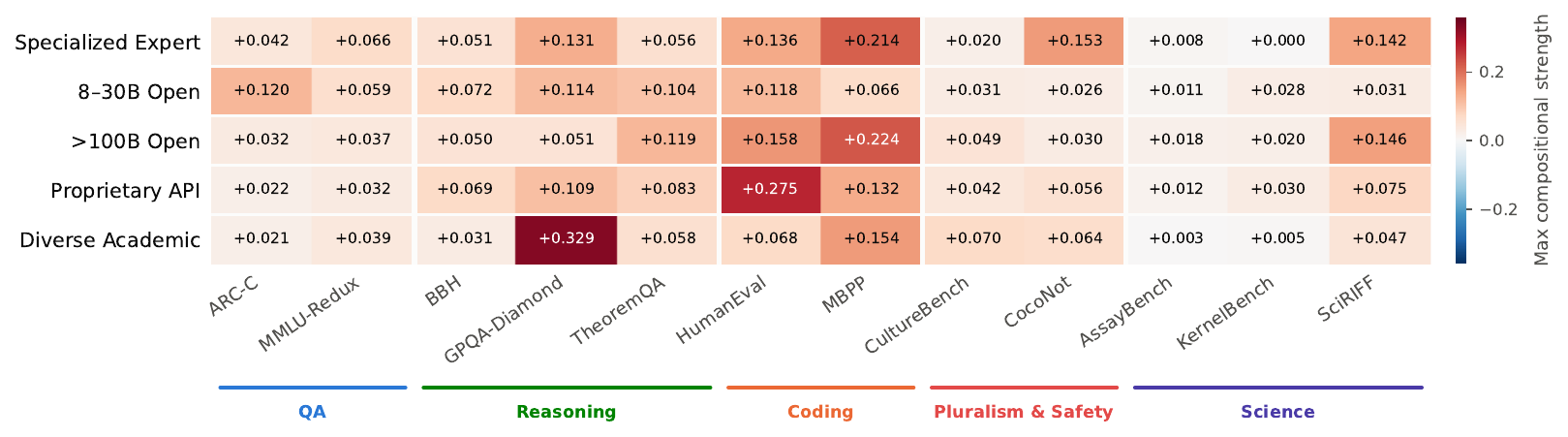}
\end{center}
\caption{Maximum compositional strength achieved by any collaboration method, for each (model pool, dataset) pair: the per-dataset counterpart to the domain-level aggregation in Figure~\ref{fig:domain-pool-heatmap}. Datasets are grouped by category (column separators and colored labels); each cell reports $\max_{m} \bar\phi(m,t)$ for dataset $t$.}
\label{fig:gain-consistency}
\end{figure}

Figure~\ref{fig:gain-consistency} refines two of the domain-level patterns from Figure~\ref{fig:domain-pool-heatmap}. First, the Coding datasets (HumanEval, MBPP) each show consistently positive gains individually, not just as a category average, confirming that Coding's strength is a property of the domain rather than an artifact of aggregating over its two datasets. Second, and in contrast to a naive reading of the domain-level heatmap, Science is \emph{not} a uniformly weak domain: it is bimodal at the dataset level. AssayBench stays near zero across every pool ($\le 0.018$), and KernelBench remains small throughout ($\le 0.030$), while SciRIFF alone reaches $0.142$ and $0.146$ for the Specialized Expert and $>$100B Open pools respectively, comparable to those pools' gains on Reasoning or Pluralism \& Safety. The domain-level Science column is therefore driven almost entirely by SciRIFF, and would be misleading if read as evidence that collaboration uniformly fails to help on scientific tasks. Finally, within Reasoning, GPQA-Diamond shows a markedly stronger and more consistent gain across pools than BBH or TheoremQA, most strikingly for Diverse Academic Models, whose domain-level Reasoning peak in Figure~\ref{fig:domain-pool-heatmap} ($+0.329$) is attributable entirely to GPQA-Diamond.

\subsection{The Shapley Value: Derivation and Axioms}
\label{app:shapley-axioms}

We assemble the grand coalition one model at a time and measure how much model $i$ adds at the point it joins. The Shapley value $\phi_i(v)$ is that amount averaged over all $n!$ orders. Since what $i$ adds depends only on which models precede it, the orders can be grouped by that preceding set $S$, giving the sum in \eqref{eq:shapley}. The term $v(S \cup \{i\}) - v(S)$ is the \emph{marginal contribution} of model $i$ to coalition $S$, and the weight $\frac{|S|!\,(n-|S|-1)!}{n!}$ is exactly the fraction of the $n!$ orderings of $\mathcal{N}$ in which all members of $S$ precede $i$ and every other model follows.

The Shapley value is the unique attribution rule satisfying four axioms \citep{shapley1953value}: the shares sum to the value the full coalition creates, interchangeable models receive equal credit, a model that never changes any outcome receives nothing, and attribution is linear when games are combined. We thus take it as the definition of compositional strength rather than one heuristic among many: under it, a model's compositional strength is its expected usefulness, averaged over every coalition it might end up in.

\subsection{Comparing Contribution Estimators}
\label{app:estimators}

Section~\ref{sec:eval-method} estimates compositional strength with the leave-one-out marginal \eqref{eq:loo} rather than the exact Shapley value \eqref{eq:shapley}, which would require all $2^n$ coalition values. On a small enough pool the exact computation is affordable, which lets us check the estimator directly. We do so on the Specialized Expert pool ($n=4$, so 16 coalitions per cell), computing the exact Shapley value, the Banzhaf value, and the leave-one-out marginal from the same cached coalition values, over 4 datasets (ARC-Challenge, BBH, HumanEval, SciRIFF) and 4 collaboration methods (Prompt Routing, Multiagent Debate, LLM-Blender, AggLM). This gives 64 (model, dataset, method) observations per estimator.

\begin{figure}[h]
\begin{center}
\includegraphics[width=\linewidth]{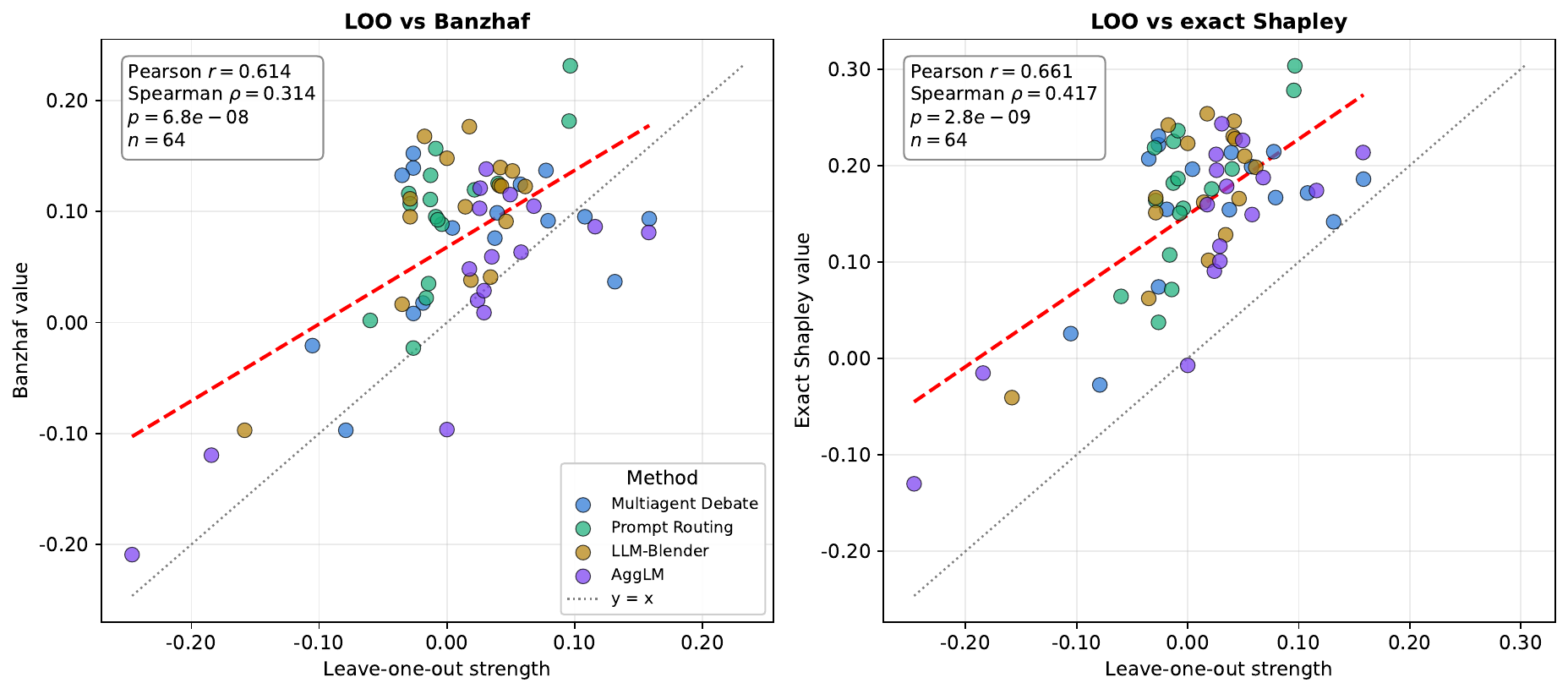}
\end{center}
\caption{Leave-one-out strength against the Banzhaf value (left) and the exact Shapley value (right) on the Specialized Expert pool. Each point is one (model, dataset, method) cell, colored by collaboration method; the dotted line is $y=x$ and the dashed line is the least-squares fit.}
\label{fig:estimator-correlation}
\end{figure}

Figure~\ref{fig:estimator-correlation} shows that the leave-one-out marginal tracks both of the more expensive estimators. The relationship is positive and highly significant in each case, with Pearson $r = 0.61$ against Banzhaf ($p = 6.8\times10^{-8}$) and $r = 0.66$ against exact Shapley ($p = 2.8\times10^{-9}$). Two qualifications are visible in the same plots. First, agreement is stronger in magnitude than in rank: the Spearman correlations are $0.31$ and $0.42$ respectively, so the estimators do not order models identically even when they broadly agree on how much each contributes. Second, nearly every point lies above the $y=x$ line, meaning the leave-one-out marginal reads systematically lower than either alternative. This is the expected direction. Measuring a model's contribution only against the full pool lets the remaining models compensate for its removal, which suppresses the credit assigned to models whose competence is duplicated elsewhere in the pool, exactly the residual term isolated in \eqref{eq:loo-decomposition}.

\section{Experiment Details}
\label{app:details}

\subsection{Evaluation Setup Details}
\label{app:eval-setup}

\begin{table}[t]
\centering
\footnotesize
\setlength{\tabcolsep}{3pt}
\begin{tabular}{@{}llp{0.60\linewidth}@{}}
\toprule
Pool & $n$ & Members \\
\midrule
  Specialized Expert & 4 & Qwen2.5-7B-Instruct, Qwen2.5-Math-7B, Qwen2.5-Coder-7B-Instruct, Qwen2.5-7B-Instruct-1M \\
  8--30B Open & 7 & Qwen3-14B, Nemotron-3-Nano-30B-A3B, gpt-oss-20b, Llama-3.1-8B-Instruct, gemma-3-12b-it, phi-4, DeepSeek-R1-Distill-Qwen-14B \\
  $>$100B Open & 4 & gpt-oss-120b, deepseek-v4-pro, kimi-k2.6, nemotron-3-ultra \\
  Proprietary API & 3 & gpt-5.5, gemini-3.5-flash, bedrock-claude-opus-4-8 \\
  Diverse Academic \citep{feng2026scalingparticipationmodularai} & 8 & parti\_0\_full, parti\_1\_full, parti\_2\_full, parti\_3\_full, parti\_4\_full, parti\_5\_full, parti\_6\_full, parti\_7\_full \\
\bottomrule
\end{tabular}
\caption{\textbf{The five model pools.} Each pool defines a separate cooperative game; compositional strengths are computed within a pool and never compared across pools. \emph{Diverse Academic} is a set of independently post-trained research models, each specialised for a different task (narrative claim verification, tool use, planner optimisation, architecture ablation, instruction following, instruction induction, conditional-probability estimation, and multi-document reasoning respectively).}
\label{tab:pools}
\end{table}

\paragraph{Model pools.} Table~\ref{tab:pools} lists the members of each of the five pools. Pools were chosen to vary along two axes that plausibly matter for collaboration, the scale of the participants and how heterogeneous their training is, ranging from four sibling 7B models differing only in their specialisation to eight independently post-trained research models with nothing in common but a rough size class.

\paragraph{Datasets.} Table~\ref{tab:datasets} lists the twelve datasets, their category, and the number of test items. Every coalition value is computed on the full test split, and every scorer is rule-based or execution-based, so no model acts as a judge anywhere in the evaluation. This matters for the leave-one-out differences in \eqref{eq:loo}: because the same items are scored the same way for every coalition, a change in $\hat{\phi}_i$ reflects the removal of model $i$ and nothing else.

\begin{table}[t]
\centering
\footnotesize
\setlength{\tabcolsep}{4pt}
\begin{tabular}{@{}llr@{}}
\toprule
Dataset & Category & Test items \\
\midrule
  ARC-Challenge \citep{clark2018think} & QA & 1172 \\
  MMLU-Redux \citep{gema2025we} & QA & 1000 \\
  GPQA-Diamond \citep{rein2024gpqa} & Reasoning & 99 \\
  BBH \citep{suzgun2023challenging} & Reasoning & 1000 \\
  TheoremQA \citep{chen2023theoremqa} & Reasoning & 400 \\
  HumanEval \citep{chen2021codex} & Coding & 114 \\
  MBPP \citep{austin2021program} & Coding & 487 \\
  CulturalBench-Hard \citep{chiu2024culturalbench} & Pluralism & 613 \\
  CocoNot \citep{brahman2024art} & Pluralism & 1000 \\
  SciRIFF \citep{wadden2025sciriff} & Science & 139 \\
  KernelBench \citep{ouyang2025kernelbench} & Science & 125 \\
  AssayBench \citep{debrouwer2026assaybench} & Science & 334 \\
\bottomrule
\end{tabular}
\caption{\textbf{The twelve evaluation datasets.} Coalition values are computed on the full test split. All scoring is rule-based or execution-based; no model acts as a judge.}
\label{tab:datasets}
\end{table}

\paragraph{Collaboration methods.} All twelve methods are run through MoCo \citep{feng2026moco} with the configurations below; unless stated otherwise generation uses temperature $0.7$ and top-$p$ $0.9$. \emph{Prompt Routing} and \emph{Graph Router} consume one natural-language description per pool member, the latter also training a small GNN router over MiniLM embeddings of the query ($8$ hidden features, $3$ in-edges, learning rate $10^{-4}$, $500$ epochs). \emph{Trained Router} LoRA-trains a Qwen2.5-7B router ($r{=}64$, $\alpha{=}16$, five epochs, learning rate $10^{-5}$) against a Skywork-Reward-Llama-3.1-8B reward model. \emph{Switch Generation} uses a fixed Qwen2.5-7B selector with a patch size of $50$ tokens. \emph{Multiagent Debate} runs three rounds; \emph{Multiagent Feedback} runs one feedback-and-refine cycle with one critic per answer; \emph{Heterogeneous Swarms} searches with a population of five for five iterations. \emph{SLM-MUX} draws three samples per model per question and breaks ties on the development split. \emph{LLM-Blender} ranks candidates listwise and fuses the top three, with both ranker and fuser fixed to an off-the-shelf Qwen2.5-7B; fixing the fuser matters for validity, since letting it default to the first pool member would silently change the fuser whenever that member is the one being left out. \emph{AggLM} GRPO-trains a 1.5B aggregator per coalition, \emph{Multiagent Finetuning} LoRA-trains generator and critic copies of every pool member, and \emph{Sparta Alignment} DPO-trains the pool over three iterations.

\subsection{C-RL Implementation Details}
\label{app:crl-impl}

\paragraph{Datasets.} Table~\ref{tab:datasets-crl} lists the ten datasets used in \S\ref{sec:train}. The eight training datasets are sized by their MoCo development split, which we further divide $8{:}2$ into training and validation for C-RL; the two datasets used only for the held-out generalization evaluation of \S\ref{sec:train-analysis} are sized by their test split instead. Two of these, KernelBench and SMDD-Bench, score the system's answer by execution rather than reference-matching. For KernelBench, the submitted CUDA kernel is compiled and benchmarked against the reference implementation on one H100 GPU: correctness is checked over randomized-input trials, and the reward is $0.25$ for a correct-but-not-faster kernel, jumping to $0.5$ for any measured speedup, with the remaining $0.5$ scaled linearly by speedup magnitude up to a difficulty-dependent cap ($10\times$/$5\times$/$2\times$ for Level 1/2/3 problems in KernelBench). For SMDD-Bench, the submitted molecule is checked with RDKit against a set of hard constraints (molecular weight, LogP, TPSA, hydrogen-bond counts, rotatable bonds, PAINS/Brenk filter matches, and Tanimoto similarity to a reference) for lead-optimization items, or scored by the recall and specificity of a submitted pharmacophore rule against held-out active and decoy molecules for pharmacophore items, depending on problem type. AssayBench, by contrast, requires no execution: the model submits a ranked list of genes, which is scored by Adjusted nDCG@100 against a precomputed relevance table from a real CRISPR genetic screen, with unmatched or hallucinated genes scored as zero relevance rather than dropped from the ranking. Whether the collaboration protocol is single-step (routing dispatches each query to one model for one generation) or multi-step (agentic and training-based methods exchange several rounds before a final answer), the execution tool is invoked as many times as the protocol's own structure requires; \eqref{eq:turn-credit}  broadcasts the final system reward score to every policy generation the rollout produced.

\begin{table}[t]
\centering
\footnotesize
\setlength{\tabcolsep}{4pt}
\begin{tabular}{@{}llrr@{}}
\toprule
Dataset & Category & Development & Test \\
\midrule
  GPQA-Diamond \citep{rein2024gpqa} & Answer-matching & 99 & 99 \\
  TheoremQA \citep{chen2023theoremqa} & Answer-matching & 400 & 400 \\
  MMLU-ProX \citep{xuan2025mmlu} & Answer-matching & 500 & 500 \\
  CulturalBench-Hard \citep{chiu2024culturalbench} & Answer-matching & 613 & 613 \\
  ChemBench \citep{Mirza2025} & Answer-matching & 498 & 500 \\
  KernelBench \citep{ouyang2025kernelbench} & Execution-verified & 125 & 125 \\
  SMDD-Bench \citep{han2026smddbench} & Execution-verified & 135 & 137 \\
  AssayBench \citep{debrouwer2026assaybench} & Reference-ranking & 218 & 334 \\
\midrule
  APEX-Shortlist \citep{dekoninck2026matharena} & Answer-matching & -- & 40 \\
  BixBench \citep{mitchener2025bixbenchcomprehensivebenchmarkllmbased} & Answer-matching & -- & 103 \\
\bottomrule
\end{tabular}
\caption{\textbf{The ten datasets used in \S\ref{sec:train}.} The eight training datasets (top) list both their MoCo development split, which we further divide $8{:}2$ into training and validation for C-RL, and their test split, reported for reference. APEX-Shortlist and BixBench (bottom) are used only for the held-out generalization evaluation of \S\ref{sec:train-analysis}, so only their test-set size is listed. \emph{Answer-matching}, \emph{execution-verified}, and \emph{reference-ranking} follow the scoring split of \S\ref{sec:train-setup}.}
\label{tab:datasets-crl}
\end{table}

\paragraph{Infrastructure.} The policy is fully fine-tuned, not adapter-tuned. Each run uses three nodes of eight H100 GPUs, with tensor parallelism $4$, pipeline parallelism $2$, and sequence parallelism enabled; rollout generation is served separately at tensor parallelism $4$. The sequence length is $16{,}384$ tokens with a response budget of $15{,}000$, and batches are formed dynamically up to $6{,}144$ tokens per GPU. A complete C-RL run takes roughly $30$ hours on this allocation.

\paragraph{Rollouts and optimization.} Each optimizer step samples $N = 8$ questions and runs $K = 8$ independent executions of the collaboration protocol per question, so a step involves $64$ full protocol executions and a variable number of policy generations depending on how often the protocol calls the policy. Advantages are computed per question over the pooled generations as in \eqref{eq:group-advantage}, and the policy is updated with a clipped policy-gradient loss at a learning rate of $10^{-6}$. We use no KL anchor to the initial policy: the coefficient is set to zero, so the only pressure on the policy is the system's reward. Training runs for up to $100$ rollout steps.

\paragraph{Attributing generations to the policy.} \S\ref{sec:train-method} requires that the loss touch only tokens the policy actually sampled, which is non-trivial because a rollout is a transcript produced by four models. We serve the policy to the collaboration protocol behind a shim that exposes an OpenAI-compatible endpoint, so the protocol calls the policy exactly as it calls any other participant and requires no modification. Every call that reaches the shim is recorded as one entry, tagged with the question and rollout it belongs to and ordered by turn, storing the prompt tokens, the sampled response tokens, and the log-probabilities under which they were sampled. Collaborator generations are issued to other endpoints and never enter this record. The training step then consumes the recorded entries directly, so the loss is applied to the policy's own tokens under their own sampling distribution, and the collaborators' text enters only as context. This is also what makes \eqref{eq:turn-credit} straightforward to implement: the scalar reward for a rollout is broadcast to every recorded entry carrying that rollout's tag.

\paragraph{Protocol configuration during training.} Collaboration runs for two rounds. LLM-Blender fuses the top $3$ candidates and ranks them listwise with an off-the-shelf gpt-oss-120b ranker, rather than pairwise, which costs one ranking call per question instead of $\binom{n}{2}$. SLM-MUX draws $3$ samples per model per question. Generation uses temperature $0.7$ and top-$p$ $0.9$ throughout, for the policy and collaborators alike.

\paragraph{Checkpoint selection.} A held-out development split, the last $20\%$ of the training pool, is reserved for model selection and never used for gradient updates. Every $3$ steps we checkpoint and score the current policy on $16$ development questions with a single rollout each, and training stops early if the development score has not improved for $25$ consecutive evaluations. We retain the three best checkpoints by development score and report the single best. All reported numbers come from that checkpoint evaluated on the full test split with one rollout per question at a fixed seed, in both individual and collaborative mode.

\paragraph{Baselines.} GRPO and DAPO use the same infrastructure, sequence budget, and learning rate, differing only in that the reward is the policy's own correctness and the rollout is a single-model generation rather than a protocol execution. Collaborative RAFT runs the full protocol with the untrained policy, keeps the transcripts whose collaborative outcome scored well, and supervised-fine-tunes the policy on its own generations from those transcripts. Prompt nudge changes no weights and prepends a description of the collaborative system to the policy's system prompt.

\end{document}